\documentclass[11pt]{paper}

\usepackage[utf8]{inputenc}
\usepackage{setspace}
\usepackage{microtype}
\usepackage{algorithm}
\usepackage{algpseudocode}
\usepackage{amsmath}
\usepackage{amsfonts}
\usepackage{amssymb}
\usepackage{amsthm}
\usepackage{booktabs}
\usepackage{url}
\usepackage{float}
\usepackage{graphicx}
\usepackage{subcaption}
\usepackage{listings}
\usepackage{xcolor}
\lstdefinelanguage{Julia}%
  {morekeywords={abstract,break,case,catch,const,continue,do,else,elseif,%
      end,export,false,for,function,immutable,import,importall,if,in,%
      macro,module,otherwise,quote,return,switch,true,try,type,typealias,%
      using,while},%
   sensitive=true,%
   alsoother={$},%
   morecomment=[l]\#,%
   morecomment=[n]{\#=}{=\#},%
   morestring=[s]{"}{"},%
   morestring=[m]{'}{'},%
}[keywords,comments,strings]%

\newcommand{\var}[1]{{\operatorname{\mathit{#1}}}}

\newtheorem{dfn}{Definition}
\newtheorem{cnj}{Conjecture}

\floatstyle{ruled}
\floatname{model}{Model}
\newfloat{model}{H}{}

\floatstyle{ruled}
\floatname{listing}{Listing}
\newfloat{listing}{ht}{}

\title{Bob and Alice Go to a Bar}
\subtitle{Reasoning About Future With Probabilistic Programs}
\author{David Tolpin \and Tomer Dobkin}
\institution{Ben-Gurion University of the Negev, Israel}
\begin{document}
\maketitle

\begin{abstract} 
It is well known that reinforcement learning can be cast as
inference in an appropriate probabilistic model. However, this
commonly involves introducing a distribution over agent
trajectories with probabilities proportional to exponentiated
rewards. In this work, we formulate reinforcement learning as
Bayesian inference without resorting to rewards, and show that
rewards are derived from agent's preferences, rather than the
other way around. We argue that agent preferences should be
specified stochastically rather than deterministically.
Reinforcement learning via inference with stochastic preferences
naturally describes agent behaviors, does not require
introducing rewards and exponential weighing of trajectories,
and allows to reason about agents using the solid foundation of
Bayesian statistics. Stochastic conditioning, a probabilistic
programming paradigm for conditioning models on distributions
rather than values, is the formalism behind agents with
probabilistic preferences.  We demonstrate realization of our
approach on case studies using both a two-agent coordinate game
and a single agent acting in a noisy environment, showing that
despite superficial differences, both cases can be modelled and
reasoned about based on the same principles.
\end{abstract}

\section{Introduction}

The `planning as inference' paradigm~\cite{WGR+11,MPT+16}
extends Bayesian inference to future observations. The agent in
the environment is modelled as a Bayesian generative model, but
the belief about the distribution of agent's actions is updated
based on future goals rather than on past facts.  This allows to
use common modelling and inference tools, notably
\textit{probabilistic programming}, to represent computer agents
and explore their behavior.  Representing agents as general
programs provides flexibility compared to restricted approaches,
such as Markov decision processes and their variants and
extensions, and allows to model a broad range of complex
behaviors in a unified and natural way.

Planning, or, more generally, reinforcement learning, as
inference models agent preferences through conditioning agents
on preferred future behaviors.  Often, the conditioning is
achieved through the Boltzmann distribution: the probability of
a realization of agent's behavior is proportional to the
exponent of the agent's expected reward. The motivation of using
the Boltzmann distribution is not clear though. A `rational'
agent should behave in a way that maximizes the agent's expected
utility, shouldn't it? One argument is that the Boltzmann
distribution models human errors and irrationality. Sometimes,
attempts are made to avoid using Boltzmann distribution by
explicitly conditioning the agent on future goals. However, such
conditioning can also lead to irrational behavior, as the famous
Newcomb's paradox~\cite{N69} suggests.

\section{The Fable: Bob and Alice Go to a Bar}

Challenges of reinforcement learning as inference can be
illustrated on the following fable~\cite{SG14}:

\begin{quote}
Bob and Alice want to meet in a bar, but Bob left his phone at
home. There are two bars, which Bob and Alice visit with
different frequencies. Which bar Bob and Alice should head if
they want to meet?
\end{quote}

Many different settings can be considered based on this story.
Bob and Alice may know each other's preferences with respect to
the bars and to the meeting with the other person, be uncertain
about the preferences, or hold wrong beliefs about the
preferences.  Their preferences may be collaborative (both want to
meet) or adversarial (Bob wants to meet Alice, but Alice avoids
Bob).  Bob may consider Alice's deliberation about Bob's
behavior, and vice versa, recursively. It turns out that all
these scenarios can be represented by a single model of
interacting agents.  However, doing this properly, from the
viewpoint of both specification and inference, requires certain
care.

\section{Model Blueprint}

Details of conditioning and inference set aside, the overall
structure of the model is more or less obvious. There are two
thought and action models, for each of the agents. The
generative models have the same structure, but apparently
different parameters (Model~\ref{mod:agent-model}).
\begin{model}
    \caption{Agent model: choosing an action based on the belief
          about preferences of the other agent}
    \label{mod:agent-model}
    \begin{algorithmic}[1]
        \Procedure{Agent}{$\theta$, $\tau'$}
            \State $\theta' \sim D_{\theta'}(\tau')$ \Comment{Draw other agent's preferences}
            \State $a \sim D_a(\theta)$ \Comment{Draw the agent's action}
            \State $\mathbb{I}^s \sim  D_s(a, \theta')$
            \Comment{Draw the action's success}
        \EndProcedure
    \end{algorithmic}
\end{model}

The parameters reflect the agent's preferences $\theta$, as
well as beliefs of the agent about the other agent's
preferences $\tau'$. The model first draws the other agent's
preferences $\theta'$ and the agent's action $a$. Then, the
model draws the action's success from a distribution
parameterized by $a$ and $\theta'$. The success
distribution $D_s$ is intentionally kept vague here, and is the
subject of the rest of the article. 

Interaction between Bob and Alice is simulated by running
inference in each of the models and then by performing an
action following (deterministically or stochastically) from
inference results. In an episodic game, the model is conditioned
on $\theta'$ and $\mathbb{I}_s$ for each agent, and the distributions of
agents' actions are inferred. Then the move is simulated by
drawing a sample from each of the posterior action
distributions (Algorithm~\ref{alg:episode-simulation}).
\begin{algorithm}
    \caption{Simulation of an episode}
    \label{alg:episode-simulation}
    \begin{algorithmic}[1]
    \State $D_{Alice} \gets$ {\sc Infer}(\Call{Agent}{$\theta_{Alice}, \tau_{Bob}$} $\vert\, \theta'=\theta_{Bob}, \mathbb{I}^s$)
    \State $D_{Bob} \gets$ {\sc Infer}(\Call{Agent}{$\theta_{Bob}, \tau_{Alice}$} $\vert\, \theta'=\theta_{Alice}, \mathbb{I}^s$)
    \State $a_{Alice} \sim D_{Alice}$
    \State $a_{Bob} \sim D_{Bob}$
    \end{algorithmic}
\end{algorithm}

Since in a single episode there is no earlier evidence about the
other agent's preferences, $\theta'$ is taken to be
known, and $\tau'$ has no effect.

Bob and Alice can even engage in a multiround game, in which
each agent updates his or her own beliefs about the other
agent's preferences based on the observed actions. To update an
agent's belief about the other agent's preferences, the model is
conditioned on an observed action $a$ and success $\mathbb{I}_s$, and
$\theta'$ is inferred, and is later used to choose an action in the
next round.

A few ways to specify the agents and perform inference were
proposed. We believe that some of them are wrong,  and others
can be streamlined. In what follows, we propose a purely
Bayesian generative approach to reasoning about future in
multi-agent environments.

\section{Background}

\subsection{Planning as Inference}

Planning, as a discipline of artificial intelligence, considers
agents acting in environments~\cite{PM17}. The agents have
beliefs about their environments and perform actions which bring
them rewards (or regrets). AI planning is concerned with
algorithms that search for policies --- mappings from 
agents' beliefs about the environment to their actions. In
planning-as-inference approach~\cite{TS06,BA09}, policy search
is expressed as inference in an appropriate probabilistic model.
The prevailing approach to casting planning as inference is
inspired by the stochastic control theory~\cite{K07} and is
based on the use of Boltzmann distribution, which ascribes to a
policy the probability proportional to the exponent of the
expected total reward~\cite{WGR+11,MPT+16}.

A probabilistic program reifying the planning-as-inference
approach encodes instantiation of policy $\pi_\theta(E)$,
depending on latent parameters $\theta$, in environment $E$. In
the course of execution, the program computes the reward $r \sim
\pi_\theta(E)$, stochastic in general.  The posterior
distribution of policy parameters $p(\theta\vert E)$, conditioned on
the environment, is defined in terms of their prior distribution
$p(\theta)$ and of the expected reward:
\begin{equation}
    p(\theta\vert E) \propto p(\theta)\exp\left(\mathbb{E}\left[r \sim \pi_\theta(E)\right]\right)
    \label{eqn:planning-as-inference}
\end{equation}
Posterior inference on \eqref{eqn:planning-as-inference} gives a
distribution of policy parameters, with the mode of the
distribution corresponding to the policy maximizing the expected
reward.

\subsection{Stochastic Conditioning}
\label{sec:stochastic-conditioning}

Stochastic conditioning~\cite{TZR+21} extends
deterministic conditioning $p(x\vert y=y_0)$,
i.e.~conditioning on some random variable in our program $y$
taking on a particular value $y_0$, to conditioning $p(x\vert y
\sim D_0)$ on $y$ having the marginal distribution $D_0$.
A probabilistic model with stochastic conditioning is a tuple
$(p(x, y), D)$ where 
\begin{itemize}
    \item $p(x, y)$ is the joint probability density of random
        variable $x$ and observation $y$, 
    \item $D$  is the distribution from which observation $y$ is
        marginally sampled, and it has a density $q(y)$.
\end{itemize}

Unlike in the usual setting, the objective is to infer $p(x\vert y
\sim D)$, the distribution of $x$ given \textit{distribution}
$D$, rather than an individual observation $y$.  To accomplish
this objective, one must be able to compute $p(x, y \sim
D)$, a possibly unnormalized density on $x$ and distribution
$D$. As usual,  $p(x, y \sim D)$ is factored as $p(x)p(y \sim D\vert x)$ where
$p(y \sim D\vert x)$ is the following unnormalized conditional
density:
\begin{equation}
    p(y \sim D\vert x) = \exp \left( \int_Y (\log p(y\vert x))\,q(y)dy \right)
    \label{eqn:prob-D-given-x}
\end{equation}

An intuition behind the definition can be seen by
rewriting~\eqref{eqn:prob-D-given-x} as a type II geometric
integral:
\begin{equation}
    \label{eqn:geometric}
    \nonumber p(y \sim D\vert x) = \prod\nolimits_Y p(y\vert x)^{q(y)dy}.
\end{equation}
Hence, \eqref{eqn:prob-D-given-x} can be interpreted as the
probability of observing \textit{all} possible draws of $y$ from
$D$, each occurring according to its probability $q(y)dy$.

By convention, in statistical notation $y \sim D$ is placed
above a rule to denote that distribution $D$ is observed through
$y$ and is otherwise unknown to the model, as 
in~\eqref{eqn:notation-sampling}. 
\begin{equation}
	\begin{aligned}
		y & \sim D \\ \midrule
		x & \sim \textit{Prior} \\
		y\vert x & \sim \textit{Conditional}(x)
	\end{aligned}
	\label{eqn:notation-sampling}
\end{equation}

\subsection{Epistemic Reasoning (Theory of Mind)}
\label{sec:epistemic-reasoning}

Epistemic reasoning~\cite{SG14,KG15,HFW+17}, also known as theory of
mind, is mutual reasoning of multiple agents about each others'
beliefs and intentions. Epistemic reasoning comes up in
the planning-as-inference paradigm when each agent's policy is
mutually conditioned on other agents' policies.

Probabilistic programming allows natural representation of
epistemic reasoning: the program modelling an agent invokes
inference on the programs modelling the rest of the agents.
However, this means that inference in multiagent settings
requires nested conditioning~\cite{R18}, which is computationally
challenging in general, although attempts are being made to find
efficient implementations of nested conditioning in certain
settings~\cite{TZR+21}.  Since each agent's model recursively
refers to models of other agents, the recursion can unroll
indefinitely.  In a basic approach~\cite{SG14,ESS+17}, the
recursion depth is bounded by a constant.

\section{Common Mistakes}

Statistical models are most suited for reasoning about the
present. Reasoning about the past or, in particular, the future
is counterintuitive and hard to get right (and should probably
be avoided when possible~\cite{T99}). Multi-agent planning as
inference, exemplified by the fable about Bob and Alice,
presents logical and statistical traps which are easy to fall
into, often without even noticing. One common mistake is
conditioning on a future event as on a present observation,
which is related to the Newcomb's paradox; however there are
also other mistakes.  Incorrect treatments of the problem in the
literature, presented in the rest of this section,  accentuate
the need for a probabilistically sound Bayesian approach to
planning as inference, the subject of this work.

\subsection{Inconsistent Preferences}

One seemingly natural way to account for agent's preferences is
to condition the agent's choices on the anticipated
event~\cite{SG14,ESS+17}.  Consider, for simplicity, one
particular variant of the fable,
in which Bob and Alice both generally choose the first bar with
probability 55\% and the second bar with probability 45\% and want to
meet~\cite[Chapter~6]{ESS+17}. The agent model (of Alice, just
to be concrete) combines the prior on Alice's behavior, which
is the Bernoulli distribution $\mathbb{I}_{Alice}^1 \sim
\mathrm{Bernoulli}(p_{Alice}^1=0.55)$, and the conditioning on Alice meeting
Bob. Alice does not know Bob's location, but can reason about
the distribution of Bob's choices $\mathrm{Bernoulli}(\hat
p_{Bob}^1)$. $\hat p_{Bob}^1$ coincides with Bob's prior
$p_{Bob}^1=0.55$ in the simplest case, but can be also
influenced by Alice's consideration of Bob's reasoning about
Alice, which is a case of epistemic reasoning
(Section~\ref{sec:epistemic-reasoning}).
\begin{model}
    \caption{Conditioning on the future as though it were the
    present.}\label{mod:future-as-present}
    \begin{algorithmic}
        \State $\mathbb{I}_{Alice}^1 \sim \mathrm{Bernoulli}(p_{Alice}^1)$
    \If {$\mathbb{I}_{Alice}^1$}
        \State $1 \sim \mathrm{Bernoulli}(\hat p_{Bob}^1)$
    \Else
        \State $0 \sim \mathrm{Bernoulli}(\hat p_{Bob}^1)$
    \EndIf
    \end{algorithmic}
\end{model}

The posterior probability of Alice going to the first bar
according to Model~\ref{mod:future-as-present}
is $\frac {0.55\hat p_{Bob}^1} {0.55\hat p_{Bob}^1 + 0.45(1-\hat p_{Bob}^1)}$. If
$\hat p_{Bob}^1 > 0.5$, that is, if Alice maintains that Bob prefers
the first bar, Alice will go to the first bar more often if she
wants to meet Bob, \textit{with the probability in which it
would meet Bob in the first bar if she would not adjust her
behavior}. It may seem (and it is indeed the course of reasoning
in~\cite{SG14} and~\cite{ESS+17}) that Alice makes a better
choice thinking about Bob, and the more she thinks about Bob
(willing to meet Alice and taking into consideration that Alice
is willing to meet Bob, and so on) the higher is the probability
of Alice to choose the first bar. 

However, on a slightly more thorough consideration, Alice's
deliberation is irrational. If Alice wants to meet Bob and knows
that Bob chooses the first bar more often, no matter to which
extent, she must always choose the first bar! Moreover, the
model's recommendation to choose the first bar with the
probability that she meets Bob in the first bar if she does
\textit{not} choose the first bar more often than usual, sounds
at least surprising and suggests that the model is wrong. 

Indeed, the above model suffers from two problems. First, it
conditions on the future as though it were the present, that is
as though Alice knew, in each case, Bob's choices. Second, the
agents' preferences with respect to the choice of a bar and to
meeting each other are expressed using different languages.
Alice chooses the first or the second bar with a non-trivial
probability, but wants to meet Bob \text{non-probabilistically}.
Moreover, it is not immediately obvious, whether and how a
probability can be ascribed to a desire (rather than a future
event). Mixing two incompatible formulations causes confusion
and gives self-contradictory results.

\subsection{Avoiding Nested Conditioning}

\cite{SMW18} approach a pursuit-evasion problem, in the form of
a chaser (sheriff) and a runner (thief) moving through city
streets, with tools of probabilistic programming and planning as
inference. A simpler but similar setting can be expressed using
Bob and Alice with adversarial preferences with respect to
meeting each other: Bob (the chaser) wants to meet Alice, but
Alice (the runner) wants to avoid Bob.

\cite{SMW18} formulates the inference problem in terms of the
Boltzmann distribution of trajectories based on the travel time
of the runner, as well as the time the runner seen by the
chaser, positive for the chaser (wants to see the runner as long
as possible) and negative for the runner (wants to avoid being
seen by the chaser as much as possible).  However, looking 
for efficient inference, this work proposes to replace nested
conditioning by conditioning of each level of reasoning on a
single sample from the previous level. This is, again,
conditioning on the future as though the future were known, and
leads to wrong results.

The problem can be hard to realize on the complicated setting
explored in the work, but becomes obvious on the example of Bob
and Alice. Consider, for simplicity, the setting in which Alice
has equal preferences regarding the bars and attributes reward 1
to avoiding Bob and 0 to meeting him. Bob chooses the first bar
in 55\% of cases; Alice employs a single level of epistemic
reasoning. It is easy to see that the optimal policy for Alice
is to always go to the second bar for the expected reward or 0.55.
However, the mode of the posterior inferred using the algorithm
in~\cite{SMW18} is for Alice to choose the second bar with
probability of 0.55, for the expected reward of 1! This is
because Alice's decision is (erroneously) conditioned on Bob's
anticipated choice of a bar, and hence Alice pretends that she
can always choose the other bar (which she cannot).

\section{Deterministic Preferences}
\label{sec:detpref}

The fable of Bob and Alice is underspecified: we know how strong
Bob or Alice prefer the first bar over the second one, or
vice versa, however we do not know how strong Bob and Alice
prefer to meet (or avoid) each other. We need a consistent
language for expressing both preferences. One common way of
expressing preferences is by ascribing rewards to action
outcomes~\cite{ESS+17}. In this way, the agent's preferences are
specified using three quantities:
\begin{enumerate}
\item $r_1$ --- the reward for visiting the first bar;
\item $r_2$ --- the reward for visiting the second bar;
\item $r_m$ --- the reward for meeting the other person.
\end{enumerate}
However, the preference of visiting the first bar in 55\% of
cases (or any other percentage different from  100\% and
0\%) cannot be expressed using rewards with a rational agent,
that is with an agent maximizing his or her expected utility. If
Alice prefers the first bar ($r_{Alice}^1 > r_{Alice}^2$) and is
rational, she should always go to the first bar.
If, in addition, Alice ascribes reward $r_{Alice}^m$ to meeting Bob,
and believes that Bob goes to the first bar with
probability $\hat p_{Bob}^1$, then she should go to the first
bar if $r_{Alice}^1 + \hat p_{Bob}^1r_{Alice}^m >
r_{Alice}^2 + (1 - \hat p_{Bob}^1)r_{Alice}^m$, and to
the second bar otherwise, ignoring ties. It is a
well-known and easy to prove fact that rational agents choose
actions deterministically.

However, stochastic behavior is not unusual, and we should be
able to model it. To reconcile rewards and stochastic action
choice, the softmax agent was proposed and is broadly used for
modelling `approximately optimal agents'~\cite{SG14}\footnote{We
delay discussion of the meaning of `approximate optimality' and
stochastic preferences till the next section.}.  A softmax agent
chooses actions stochastically, with probabilities proportional
to the exponentiated action utilities. If Bob or Alice choose a
bar without caring about meeting the other party, the action
utility is the reward for visiting a bar, so to model e.g.
Alice choosing the first bar over the second one with
probability $p_{Alice}^1$, we should ascribe rewards
$r_{Alice}^1$ an $r_{Alice}^2$ such that $r_{Alice}^1 =
r_{Alice}^2 + \log\left(\frac {p_{Alice}^1} {1 -
p_{Alice}^1}\right)$. Concretely, for $p_{Alice}^1 = 0.55$ we
obtain $r_1 \approx r_2 + 0.2$; only the difference between
rewards matters in a softmax agent, rather than their absolute
values. 

If, however, Alice's willingness to meet (or avoid) Bob is
considered, the utilities of Alice's actions are updated based
on Alice's belief that Bob goes to the first bar:
\begin{equation}
    \begin{aligned}
        u_{Alice}^1 & = r_{Alice}^1 + \hat p_{Bob}^1r_{Alice}^m \\
        u_{Alice}^2 & = r_{Alice}^2 + (1 - \hat p_{Bob}^1)r_{Alice}^m
    \end{aligned}
    \label{eqn:softmax-u-alice}
\end{equation}
and Alice chooses the first bar with probability
\begin{equation}
\Pr(\mathbb{I}_{Alice}^1) = \frac {\exp(u_{Alice}^1)} {\exp(u_{Alice}^1) + \exp(u_{Alice}^2)}
\label{eqn:softmax-a-alice}
\end{equation}
Alice's behavior is stochastic, and depends on Alice's belief
$\hat p_{Bob}^1$ about Bob's, similarly stochastic, behavior.

\section{Probabilistic Preferences}
\label{sec:probabilistic-preferences}

Section~\ref{sec:detpref} shows how stochastic behavior can be
modelled using a softmax agent exhibiting an `approximately
optimal' behavior, that is, selecting actions with
log-probabilities proportional to their expected utilities.
However, one may wonder what would be the optimal behavior if
the behavior we model with softmax is only `approximately
optimal' and hence suboptimal. We believe that treating
stochastic behavior as suboptimal because it does not
maximize the utility based on rewards we ascribe is
contradictory. When Alice chooses the first bar 55\% of
time and the second bar 45\% of time it is not because
she cannot choose better (that is, always the first bar). It is
her choice to sometimes go for a drink and have a lot of people
around and a loud music, sometimes go for a drink and have fewer
people around and read a book without disturbing music in the
background. Bob, in addition to going to one of the two bars to
enjoy some drinks and see Alice, most probably eats breakfast
every morning.  Let us speculate that sometimes Bob prefers
scrambled eggs and sometimes porridge, randomly with probability
about 0.6 towards scrambled eggs. However, if Bob had to eat
scrambled eggs every morning and to give up on porridge entirely,
because this is our perception of the optimal behavior based on
rewards we ourselves introduced, rather arbitrarily, he would
most probably be less happy about his diet. 

Assuming that there is no shortage of either oat or eggs, and
that Bob's income essentially frees him from financial
considerations in choosing one meal over the other, we should
perceive Bob's behavior with respect to either the breakfast
menu or the choice of a bar as optimal, and suggest means for
specifying optimal stochastic behaviors and reasoning about
them. One option is to postulate that softmax action selection
is indeed optimal.  However, this option involves an assumption
that softmax is a law of nature, and that there are latent
rewards which we can only observe as choice probabilities
through the softmax distribution. We believe that a separate
theory of softmax-optimal stochastic behavior is unnecessary,
and one can reason about agent behavior, both in single-agent
and multi-agent setting, using Bayesian probabilities only. The
rest of this section is an elaboration of this attitude.

Let us return to the fable. According to the fable, the optimal
probability of Alice's choice with respect to the bars is given
(observed with high confidence or proclaimed by an oracle). To
complete the formulation, it remains to specify the optimal
probability of Alice to meet Bob. The question is: probability of
what event reflects the preference?  It turns out that we need
to specify the probability of Alice choosing to visit the bar
where Bob is heading, provided  Alice knows Bob's plans and
would otherwise visit either bar with equal probability. Indeed,
this results in the following generative model:
\begin{model}
    \caption{Generative model with probabilistic
    preferences}\label{mod:detpref-gen}
    \begin{algorithmic}
        \State $\mathbb{I}_{Alice}^1 \sim Bernoulli(p_{Alice}^1)$ \label{mod:detpref-gen-prior}
        \If {$\mathbb{I}_{Alice}^1$}
            \State $\mathbb{I}_{Bob}^1 \sim Bernoulli(p_{Alice}^m)$
        \Else
            \State $\mathbb{I}_{Bob}^1 \sim Bernoulli(1 - p_{Alice}^m)$
        \EndIf
    \end{algorithmic}
\end{model}

The model, in a superficially confusing way, generates Bob's
location $\mathbb{I}_{Bob}^1$ based on Alice's desire to meet
Bob. However, the confusion can be resolved by the following
interpretation: when Alice chooses to go to the first bar, it is
because, in addition to the slight preference over the second
bar (line~\ref{mod:detpref-gen-prior}), she believes that Bob
will also go to the first bar with probability $p_{Alice}^m$. It
is Alice's anticipations of Bob's behavior that are generated by
the model, rather than events.

Since Model~\ref{mod:detpref-gen} generates
anticipations, it should also be conditioned on anticipations.
Let us, at this point, accept as a fact that Alice concludes
that Bob goes to the first bar with probability $\hat
p_{Bob}^1$. Then, the model must be conditioned on $\mathbb{I}_{Bob}^1$
\textbf{being distributed} as $\mathrm{Bernoulli}(\hat
p_{Bob}^1)$.  This involves an application of \textit{stochastic
conditioning} (Section~\ref{sec:stochastic-conditioning}):
\begin{model}
    \caption{Stochastically conditioned Model~\ref{mod:detpref-gen}}\label{mod:detpref-gen-stocond}
    \begin{algorithmic}
        \State $\mathbb{I}_{Bob}^1 \sim \mathrm{Bernoulli}(\hat p_{Bob}^1)$
        \State \rule{0.5\linewidth}{0.5pt}
        \State $\mathbb{I}_{Alice}^1 \sim \mathrm{Bernoulli}(p_{Alice}^1)$
        \If {$\mathbb{I}_{Alice}^1$}
            \State $\mathbb{I}_{Bob}^1 \sim \mathrm{Bernoulli}(p_{Alice}^m)$
        \Else
            \State $\mathbb{I}_{Bob}^1 \sim \mathrm{Bernoulli}(1 - p_{Alice}^m)$
        \EndIf
    \end{algorithmic}
\end{model}

A connection between probabilistic preferences and deterministic
preferences along with the Boltzmann distribution is revealed by
writing down the analytical form of the probability distribution
specified by Model~\ref{mod:detpref-gen-stocond}:
\begin{equation}
    \begin{aligned}
    \log & \Pr[\mathbb{I}_{Alice}^1] = \\
        & \begin{cases}
         \log p_{Alice}^1 + \hat p_{Bob}^1\log p_{Alice}^m + (1 - \hat p_{Bob}^1)\log (1 - p_{Alice}^m) & \text{if }\mathbb{I}_{Alice}^1 \\
         \log (1 - p_{Alice}^1) + \hat p_{Bob}^1\log (1 - p_{Alice}^m) + (1 - \hat p_{Bob}^1)\log p_{Alice}^m & \text{otherwise}
     \end{cases}
     \end{aligned}
     \label{eqn:log-pr-detpref-gen-stocond}
\end{equation}
That is, according to Model~\ref{mod:detpref-gen-stocond},
Alice's actions are Boltzmann-distributed with rewards
\begin{equation}
    \begin{aligned}
        r_{Alice}^1 - r_{Alice}^2 & = \log \frac {p_{Alice}^1} {1 - p_{Alice}^1} \\
        r_{Alice}^m & = \log \frac {p_{Alice}^m} {1 - p_{Alice}^m}
    \end{aligned}
    \label{eqn:detpref-rewards}
\end{equation}
A direct corollary from \eqref{eqn:log-pr-detpref-gen-stocond}
and \eqref{eqn:detpref-rewards} is that planning as inference
with probabilistic preferences can be encoded as a Bayesian
generative model using probabilities only, without resort to
rewards. However, a more general conjecture can be made:

\begin{cnj}
Preferences are naturally probabilistic in general and should be
expressed as probabilities of choices under clairvoyance. 
`Deterministic' specification of preferences through rewards is
an indirect way to encode probabilistic preferences. Reward
maximization algorithms for policy search are in fact
\textit{maximum a posteriori} approximations of the
distributions of optimal behaviors, which are inherently
stochastic.
\label{cnj:probabilistic-preferences}
\end{cnj}

Conjecture~\ref{cnj:probabilistic-preferences} may raise a
question of the role of reward maximization in cases where
rewards are given, such as games or trade. However, rewards in
games or trade are themselves invented by humans because of
difficulty apprehending probabilistic preferences. Once the
essence of probabilistic preferences is understood, settings in
which reward maximization is the optimal strategy lose their
importance as general models and become mere edge cases.

\section{Reinforcement Learning as Inference}
\label{sec:ri-as-inference}

Let us map certain well-known research problems of reinforcement
learning to Bayesian inference. Some of the problems are perceived as
having paradoxical properties, however their paradoxicality is
resolved by reformulation in terms of probabilities, without
recourse to artificially introduced rewards and utilities.

In the rest of the section, we provide an alternative, purely
probabilistic, definition of the model known as Markov decision
process (including the `partially observable' variant). Then, we
formalize the tasks of planning and apprenticeship learning.
There is an ambiguity in reinforcement learning literature with
regard to notions of reinforcement learning, inverse
reinforcement learning, apprenticeship learning, and planning. In
this work, we chose to use the term \textit{planning} to denote
inferring the (optimal or rational) agent behavior given the
model, and \textit{apprenticeship learning} to denote inferring
the model parameters given observed agent behavior.  Note that
the notion of 'optimal' behavior is different here from the
behavior considered optimal in traditional reinforcement
learning literature: rather than corresponding to the mode of
the posterior distribution of behaviors specified by the model,
it is the posterior distribution itself.

\subsection{Markov Decision Process}
\label{sec:mdp}

Customarily, a Markov decision process (MDP)~\cite{S10} is defined by
tuple $(S,A,T,\gamma,D,R)$, where $S$ is a set of states, $A$ is
a set of actions, $T$ is a state transition distribution,
$\gamma \in [0,1)$ is a discount factor, and $R$ is the reward
function. A policy $\pi$ can be imposed on a Markov decision
process, and constitutes a mapping from states to actions.
Realization of $\pi$ produces samples of agent behaviors, or
trajectories, through the state-and-action space of the MDP.
A choice of policy depends on rewards $R$, an `optimal' policy is
usually taken to maximize the expected discounted (by $\gamma$)
reward.  On the other hand, the rewards may be unknown, and
inferred from observations of the agent
behavior~\cite{NR00,AN04,HGE+17}.

The above definition of MDP provides a fertile ground  for
reinforcement learning research, however it mixes the agent, the
environment, the goal of the agent in the environment, and the
issues of practical computability, all in a single definition. In
particular, the space of states $S$ and actions $A$ describe the
agent (where it is and what it can do), the transition
probabilities $T$ is a property of the environment, and the reward
function $R$ defines the goal of the agent in the environment.
The discount factor $\gamma$ modifies the environment in
such a way that the length of the episode trajectory is
geometrically distributed, through implicitly modifying $S$ and
$T$ such that $S$ includes a terminal state and the transition
matrix includes transition from any state to the terminal state with
probability $1-\gamma$ --- and that ensures that the policy
value is bounded. The geometrical distribution is not the only
and not necessarily the most justified choice for trajectory
length distribution. The Poisson or the negative-binomial
distribution may, for example, be used instead when appropriate
(for example, the negative-binomial distribution is a natural
model for an agent repeatedly performing an action, with a
certain number of failed attempts allowed). However,
conventional MDP does not provide for such flexibility.

Bayesian modelling allows instead to define a Markov decision
process such that the agent, the environment, the agent
preferences, and the computability issues are explicit and
clearly separated. Before we proceed to our definition, let us draw
connection between the fable of Bob and Alice and a single-agent
decision process. It is relatively easy to see how
the case of two agents can be extended to a greater number of
agents. However, a two-agent setup can as well be used naturally
for modelling a single agent acting in a stochastic environment
--- by representing the stochasticity of the environment as the
other, neutral (neither adversarial nor collaborative) agent,
acting regardless of the state of the first agent. We will stick
to this paradigm here and define a 2-agent Markov
decision process, with a single-agent Markov decision process as
a special case of interaction of two agents in an environment
with the other agent being neutral.

\sloppy
\begin{dfn}A 2-agent Markov decision process (MDP) is a tuple
	$(S, A, A_1, A_2, \circ, t, m_1, m_2)$ where
	\begin{itemize}
		\item $S$ is the set of states;
		\item $A$ is the set of actions of the MDP;
		\item $A_1$,$A_2$ are the sets of actions of each agent;
		\item $\circ: A_1 \times A_2 \to A$ is the action composition operator;
		\item $t: S \times A \to S$ is the transition 
			  function of the MDP;
		\item $m_1$, $m_2$ are the agent models.
	\end{itemize}
\end{dfn}

An MDP invokes stochastic agents represented by their models $m_1$ and $m_2$
(Algorithm~\ref{alg:mdp}).  
\begin{algorithm}
    \caption{2-agent Markov decision process}
    \label{alg:mdp}
    \begin{algorithmic}[1]
		\State $s \gets s_0$
		\While {$s \ne \bot$}
			\State $a_1 \sim m_1(s)$
			\State $a_2 \sim m_2(s)$ \Comment{$m_2(s, a_1)$ in a sequential setting}
			\State $a \gets a_1 \circ a_2$
			\State $s \gets t(s, a)$
		\EndWhile
	\end{algorithmic}
\end{algorithm}
Action $a$ applied to the state is a composition $a_1 \circ a_2$
of actions $a_1$ and $a_2$ chosen by each agent. How exactly
the composition is accomplished is specific to a particular
MDP instance. An MDP episode terminates when the terminal state
$\bot$ is reached.  Conditions for the finiteness of the episode
trajectory in expectation are researched and known in the
probabilistic programming literature~\cite{MPY+18}, where the
program trace must be finite in expectation for a probabilistic
program to specify a distribution. Let us define the notion of
`episode trajectory' though:
\begin{dfn}An episode trajectory of a given MDP is a tuple
$(s_0, \pmb{a})$ where $s_0$ is the initial state and $\pmb{a}$
is the sequence of MDP actions taken until the terminal state
is reached.
\end{dfn}
An episode trajectory fully defines an episode --- given
the initial state and the actions, the intermediate states are
deterministic, according to Algorithm~\ref{alg:mdp}.

While Algorithm~\ref{alg:mdp} encodes actual unrolling of an
MDP, an agent model represents reasoning behind action
selection by the agent. Both agent models have the same
structure, and define each agent's preferences and beliefs about
the other agent. 

\begin{dfn}A 2-agent MDP agent is a model
	\begin{equation}
	a_a \sim m_a(s) = \left[
		\begin{aligned}
			a_b & \sim \widehat {D_b}(s, a_a) \\ \midrule
			a_a & \sim D_a(s) \\
			a & \gets a_a \circ a_b \\
			s' & \gets t(s, a) \\
			s' & \sim D_s(s) \\
			s' & \ne \bot \land m_a(s')
		\end{aligned}
		\right]
		\label{eqn:agent}
	\end{equation}
	where
	\begin{itemize}
		\item $a, b$ is either $1, 2$ for $m_1$ or $2, 1$ for $m_2$;
		\item $s \in S$ is the current state;
		\item $D_a: S \to A_a$ is the prior distribution of the
			agent's actions in state $s$;
		\item $\widehat {D_b}: S \times A_a \cup \{ \bot \} \to A_b$ is the agent's belief about the
			distribution of the other agent's actions in state
			$s$ given that action $a_a$ is taken;
		\item $D_s: S \to S$ is the distribution of states to which the
			agent desires to pass from state $s$.
	\end{itemize}
	\label{dfn:agent}
\end{dfn}

The rule between the first line and the rest of the
model means that the model is stochastically conditioned on the
distribution of $a_b$ (see
Section~\ref{sec:stochastic-conditioning}). The last line of the
model states that unless $s'$ is the terminal state, the model
is recursively conditioned in state $s'$.

Note that $\widehat {D_b}$  depends on both $s$ and $a_a$ in
Definition~\ref{dfn:agent}. This
addresses a sequential setting, in which agent $b$ chooses an
action after observing the action of agent $a$. In a
simultaneous setting, neither agent knows the other agent's
choice, and the choice of an action depends on the current state
only (formally, $\bot$ is passed as the second argument of
$\widehat {D_b}$).

In a partially observable MDP (POMDP), the states are not known
to the agents, but can be observed with uncertainty. From the
Bayesian modelling point of view though, this does not affect
the agent model; the only difference between MDP and POMDP here
is that in POMDP the state is a latent, rather than an
observable, variable. A proper name for POMDP from the Bayesian
perspective would be \textit{latent variable} MDP.

\subsection{Planning}

Conventional planning consists in finding a policy that
maximizes the expected total (discounted) reward. In the
probabilistic formulation of Section~\ref{sec:mdp}, a policy of
an agent that maximizes the expected total reward would correspond to
marginal, with respect to the other agent,  maximum \textit{a
posteriori} (MMAP) trajectories. Obviously, MMAP trajectories
are just an approximation of the behavior specified by the
agent, reasonably good if the distribution is sharply peaked
around the MMAP, but can be arbitrarily bad in general, as the
eggs or porridge for breakfast dilemma
(Section~\ref{sec:probabilistic-preferences}) demonstrates.

The rational behavior of the agent is specified by the agent
model~\eqref{eqn:agent}, and is in general stochastic (a
distribution of actions in each state). From the Bayesian
probabilistic standpoint, planning is just instantiation of the
distribution specified by the agent model. Common inference
algorithms, such as Markov chain Monte Carlo methods or
variational inference, can be used for representing or
approximating the policy --- the posterior distribution
of agent's actions. 

In the case of a neutral second agent, such that representing the
environment noise, posterior inference is straightforward.
However, if the agents are either collaborative or
adversarial, epistemic reasoning must be taken into account,
resulting in potentially unbounded nesting of inference.

\subsection{Apprenticeship Learning}

Apprenticeship learning~\cite{AN04} is concerned with a
situation in which the agent preferences are not given
explicitly, however there are observed MDP trajectories, from
which the preferences can be inferred. Specifically, inverse
reinforcement learning~\cite{NR00} aims at learning the reward
function of a conventional MDP given observed trajectories. A
famous paradox of inverse reinforcement learning is that many
reward functions may result in the same optimal policy, hence
the problem of learning the reward function is unidentified
\textit{per se}.

Within the Bayesian probabilistic framework, apprenticeship
learning reduces to just swapping some latent and observed
variables in the agent model~\ref{eqn:agent}. Agent preferences
are specified by $D_s$. Inference on essentially the same model
$\overline m_a$, with a slight modification that $D_s$ is a
random variable drawn from a prior, conditioned on MDP
trajectories, reifies apprenticeship learning
(Definition~\ref{dfn:apprentice}). Note that $a_a$ and $a_b$ are
observed rather than drawn here.

\begin{dfn}A 2-agent MDP apprentice model is a model
	\begin{equation}
	D_s \sim \overline{m}_a(s) = \left[
		\begin{aligned}
			D_s & \sim H_s \\
			a_a & \sim D_a(s) \\
			a & \gets a_a \circ a_b \\
			s' & \gets t(s, a) \\
			s' & \sim D_s(s) \\
			s' & \ne \bot \land \overline{m}_a(s')
		\end{aligned}
		\right]
		\label{eqn:agent}
	\end{equation}
	conditioned on an MDP trajectory, where $H_s$ is a prior
	distribution of distributions of agent preferences, and the
	rest is as in Definition~\ref{dfn:agent}.
	\label{dfn:apprentice}
\end{dfn}

\section{Numerical Experiments}

We illustrate reasoning about future with probabilistic
preferences using implementations of the fable of Bob and Alice
and of the sailing problem~\cite{PG04,KS06,TS12}.

\subsection{The Fable of Bob and Alice}

We implemented the model and inference in Gen~\cite{CSL+19}. The
model, along with supporting data types and functions, is shown
in Listing~\ref{lst:bob-alice-model}.  Since Gen does not have
built-in support for stochastic conditioning, we emulate
stochastic conditioning using Bernoulli distribution (line 29).
This is of course specific to the simple case of Bob and Alice,
and is not possible in general.

The agent model, along with supporting data types
and functions, is shown in Listing~\ref{lst:bob-alice-model}.

\begin{listing}
    \caption{Model of Bob or Alice in Gen}\label{lst:bob-alice-model}
    \begin{lstlisting}[language=Julia]
"Agent parameters"
struct Agent
    name::String
    p1::Real
    pm::Real
end

"Computes conditional log-probability of agent's choice"
function logp(a::Agent, q::Real, I1::Bool)
    if I1
        q * log(a.pm) + (1 - q)*log(1 - a.pm)
    else
        q * log(1 - a.pm) + (1 - q)*log(a.pm)
    end
end

"Defines an agent reasoning about the other agent
 with a given recursion depth"
@gen function model(me::Agent, buddy::Agent,
                    depth::Int, niter::Int)
    if depth > 0
        J1s = run_model(buddy, me, depth-1, niter)
        q = sum(J1s)/length(J1s)
    else
        q = 0.5
    end
    I1 = @trace(bernoulli(me.p1), :I1)
    # Stochastic conditioning, emulated
    @trace(bernoulli(exp(logp(me, q, I1))), :logp)
end
    \end{lstlisting}
\end{listing}

Gen requires writing the inference code, shown in
Listing~\ref{lst:bob-alice-inference}. Conditioning on
\texttt{:logp} equal to \texttt{true} refers to emulation of
stochastic conditioning in the model.
\begin{listing}
    \caption{Inference on the model in Listing~\ref{lst:bob-alice-model}}
    \label{lst:bob-alice-inference}
    \begin{lstlisting}[language=Julia]
@memoize function run_model(me::Agent, buddy::Agent,
                            depth::Int, niter=1000)
    observations = Gen.choicemap()
    observations[:logp] = true

    trace, _ = Gen.generate(model,
                            (me, buddy, depth, niter),
                            observations)
    J1s = []
    for i = 1:niter
        trace, _ = Gen.mh(trace, select(:I1))
        push!(J1s, get_choices(trace)[:I1])
    end
    return J1s
end
    \end{lstlisting}
\end{listing}

The fable of Bob and Alice is simple enough to analyze using an
analytical model, without resorting to approximate inference.
Listing~\ref{lst:bob-alice-analytical} shows the agent
implemented analytically. The agent returns the exact rather than
Monte-Carlo approximated choice distribution.
\begin{listing}
	\caption{Analytical model}\label{lst:bob-alice-analytical}
	\begin{lstlisting}[language=Julia]
"Defines an analytical model of the agent with choice
probabilities computed in closed form, rather than
through Monte Carlo inference."
@memoize function anamodel(me::Agent, buddy::Agent,
                           depth=0)
    p = me.p1
    if depth > 0
        q = anamodel(buddy, me, depth-1)
        u1 = log(p) + logp(me, q, true)
        u2 = log(1-p) + logp(me, q, false)
        p = exp(u1)/(exp(u1) + exp(u2))
    end
    p
end
	\end{lstlisting}
\end{listing}

In the experiments below, we show results obtained both by
Monte-Carlo inference on the Gen model and using the analytical
model, both to demonstrate that the Gen model, more complicated
but generalizable to a wide range of problems, returns
essentially the same results as the analytical model and to help
the reader discern easily between actual trends and
approximation errors in inference outcomes. We consider several
cases of Bob's and Alice's preferences. We run Monte Carlo
inference for 5000 iterations with 10\% burn-in. Each plot shows
the posterior choice distributions of each of the agents for a
range of deliberation depths, with depth 0 corresponding to the prior
behavior, depth 1 to the posterior using the prior behavior of
the other agent, and so on. The source code of the studies is
available at \url{https://bitbucket.org/dtolpin/playermodels/}.

\subsubsection{Bob and Alice Want to Meet}

\begin{figure}[h]
    \begin{subfigure}{0.495\linewidth}
        \includegraphics[width=\linewidth]{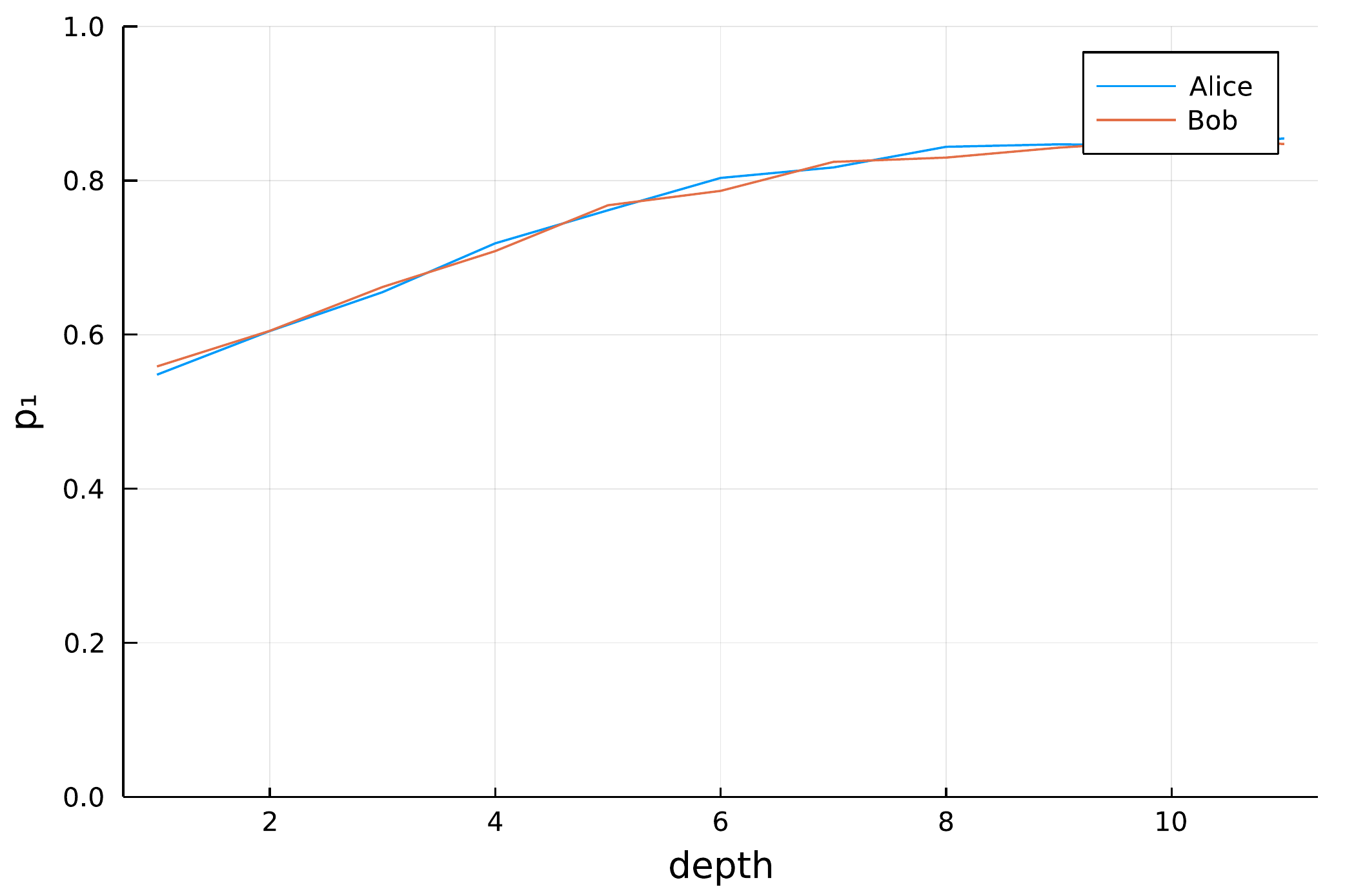}
        \caption{Monte Carlo approximation}
    \end{subfigure}
    \begin{subfigure}{0.495\linewidth}
        \includegraphics[width=\linewidth]{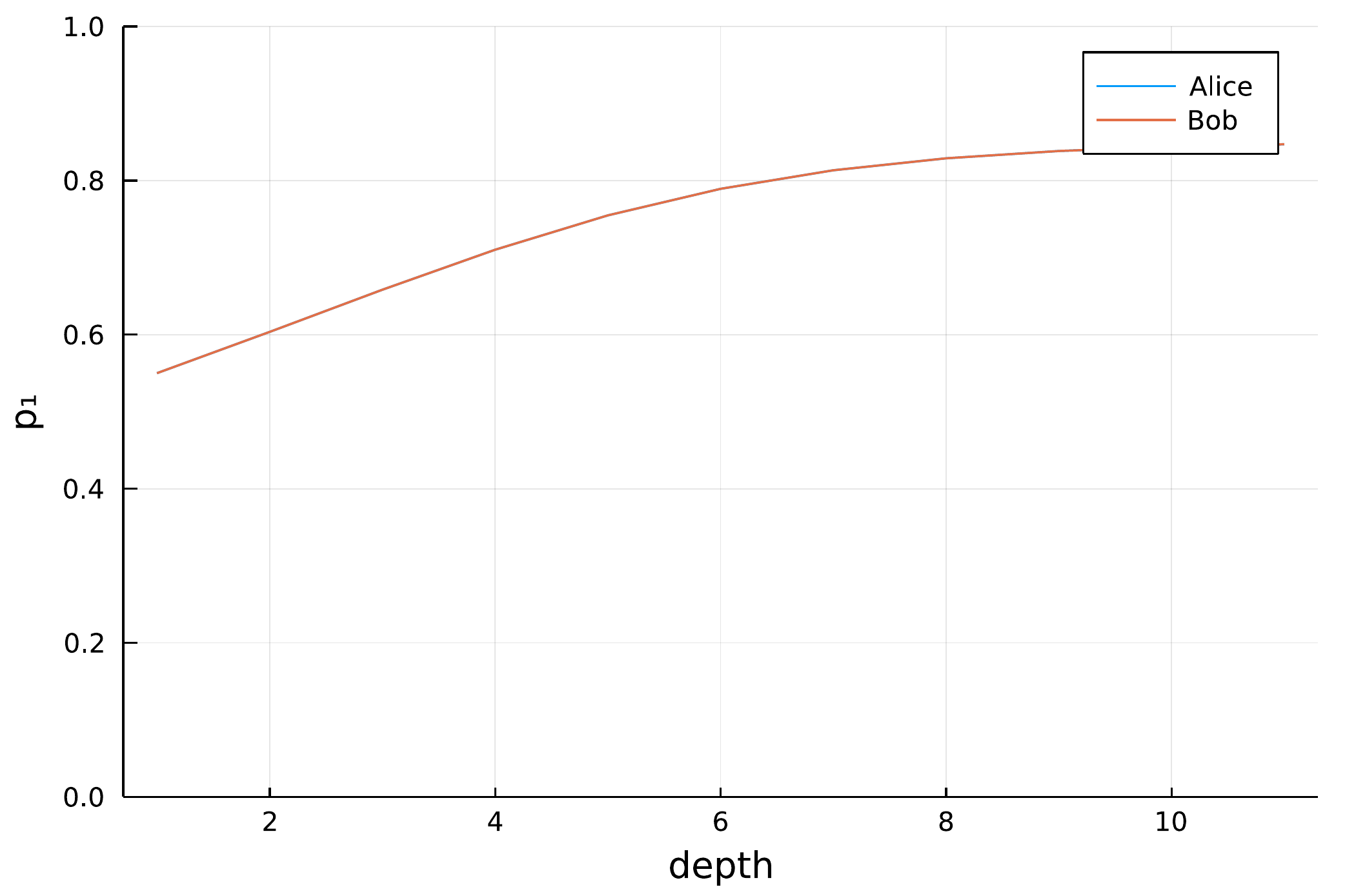}
        \caption{Analytical solution}
    \end{subfigure}
    \caption{Bob and Alice like the same bar: $p_{Alice}^1=p_{Bob}^1=0.55$, $p_{Alice}^m=p_{Bob}^m=0.9$}
    \label{fig:ba-0.55-0.9}
\end{figure}

We first explore rational behavior of Bob and Alice when they
want to meet each other. Figure~\ref{fig:ba-0.55-0.9} shows the
case when Bob and Alice have the same preferences with respect
to the bar and to meeting the other person. Posterior
choice distributions of both agents are identical, and approach
asymptotically a limit probability of going to the first bar,
which is still smaller than 1, as the depth goes to infinity.

\begin{figure}[h]
    \begin{subfigure}{0.495\linewidth}
        \includegraphics[width=\linewidth]{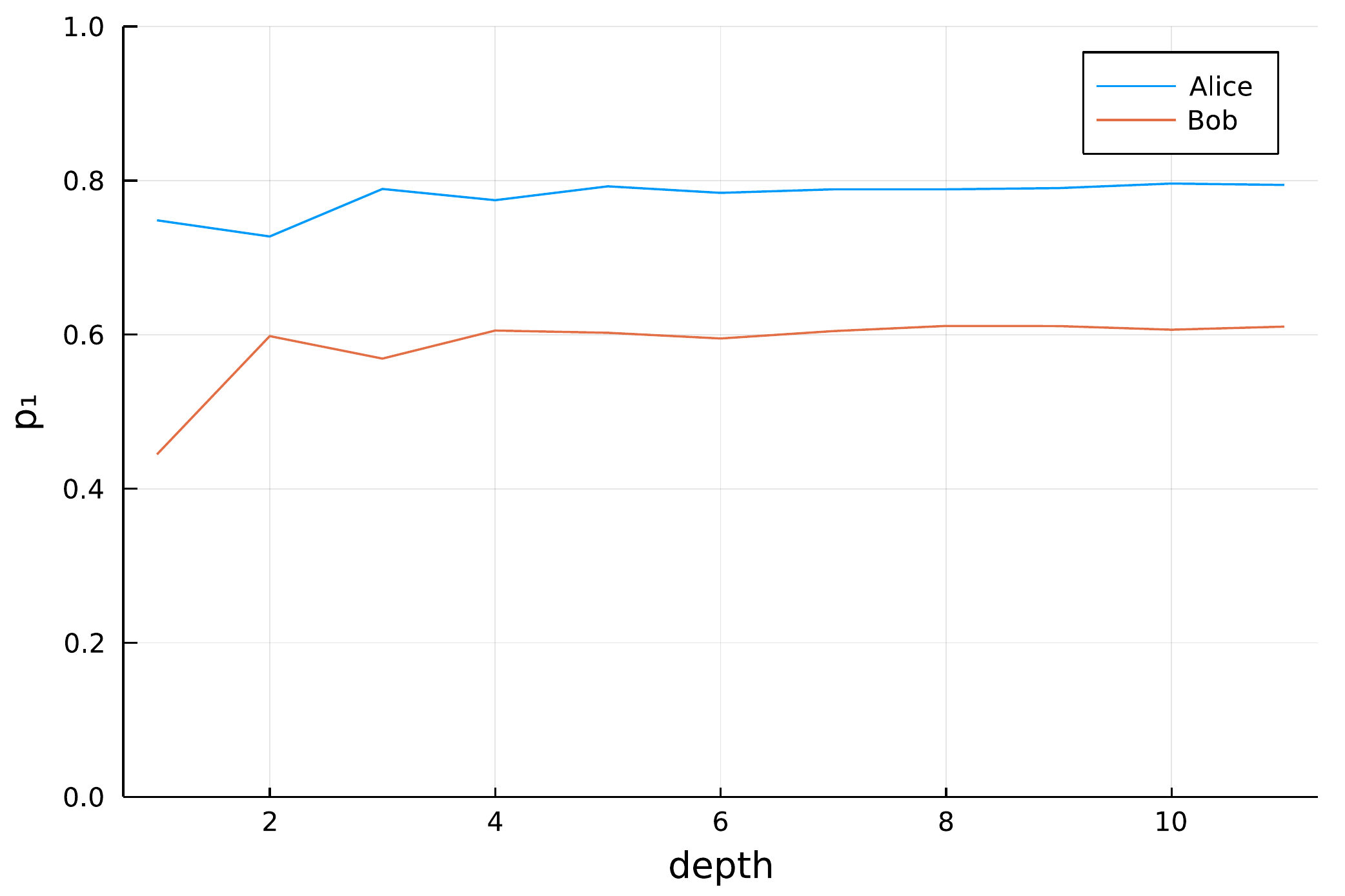}
        \caption{Monte Carlo approximation}
    \end{subfigure}
    \begin{subfigure}{0.495\linewidth}
        \includegraphics[width=\linewidth]{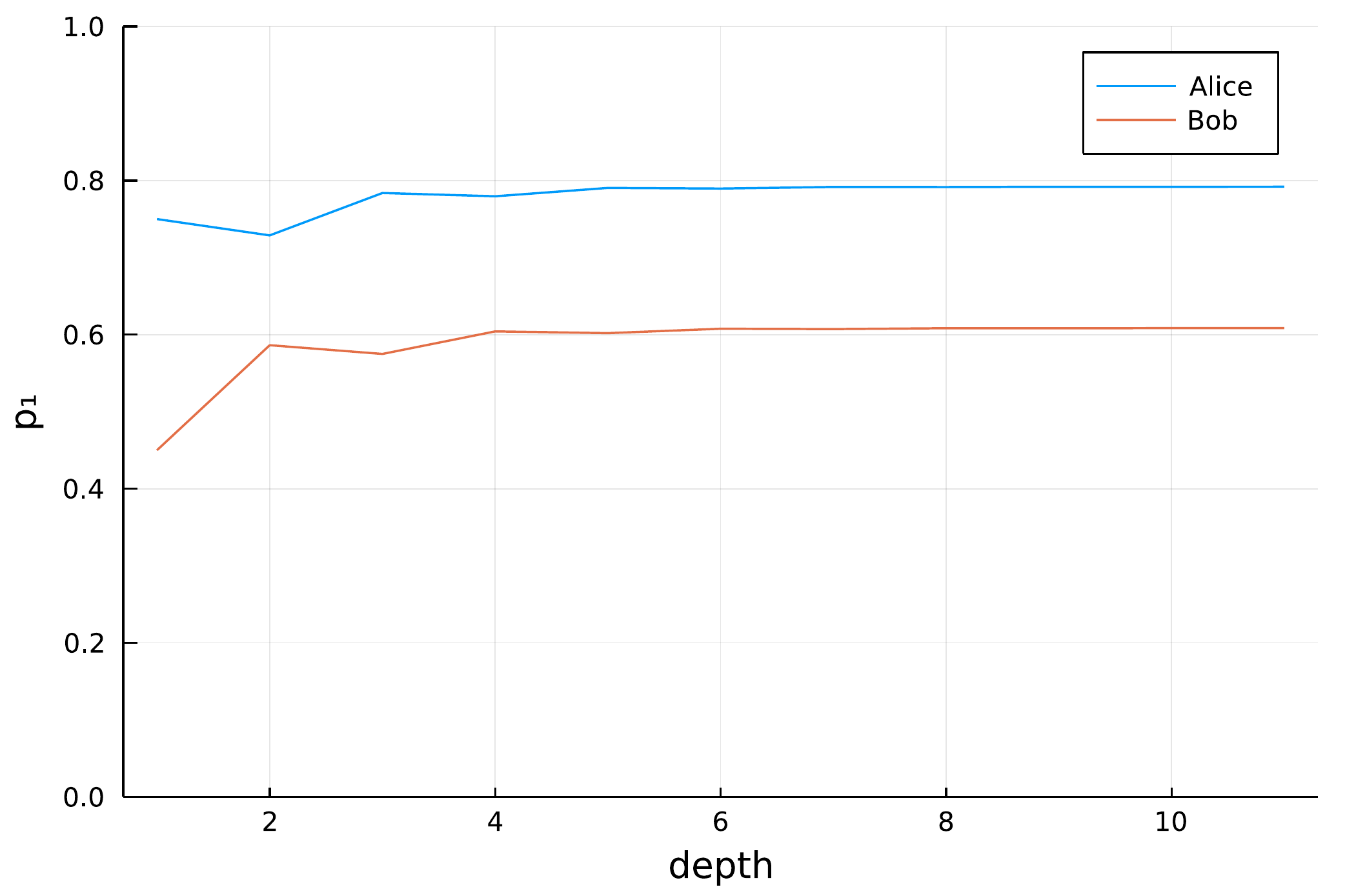}
        \caption{Analytical solution}
    \end{subfigure}
    \caption{Bob and Alice like different bars:
    $p_{Alice}^1=0.75$, $p_{Bob}^1=0.45$, $p_{Alice}^m=p_{Bob}^m=0.75$}
    \label{fig:ba-0.75-0.75-0.45-0.75}
\end{figure}

Figure~\ref{fig:ba-0.75-0.75-0.45-0.75} addresses contradictory
preferences with respect to bars --- Alice strongly prefers the
first bar, while Bob slightly prefers the second bar. Since they
still want to meet, they both choose, when the deliberation
depth is sufficient, the first bar more often than the second
one.

\subsubsection{Bob Chases Alice, Alice Avoids Bob}

\begin{figure}[h]
    \begin{subfigure}{0.495\linewidth}
        \includegraphics[width=\linewidth]{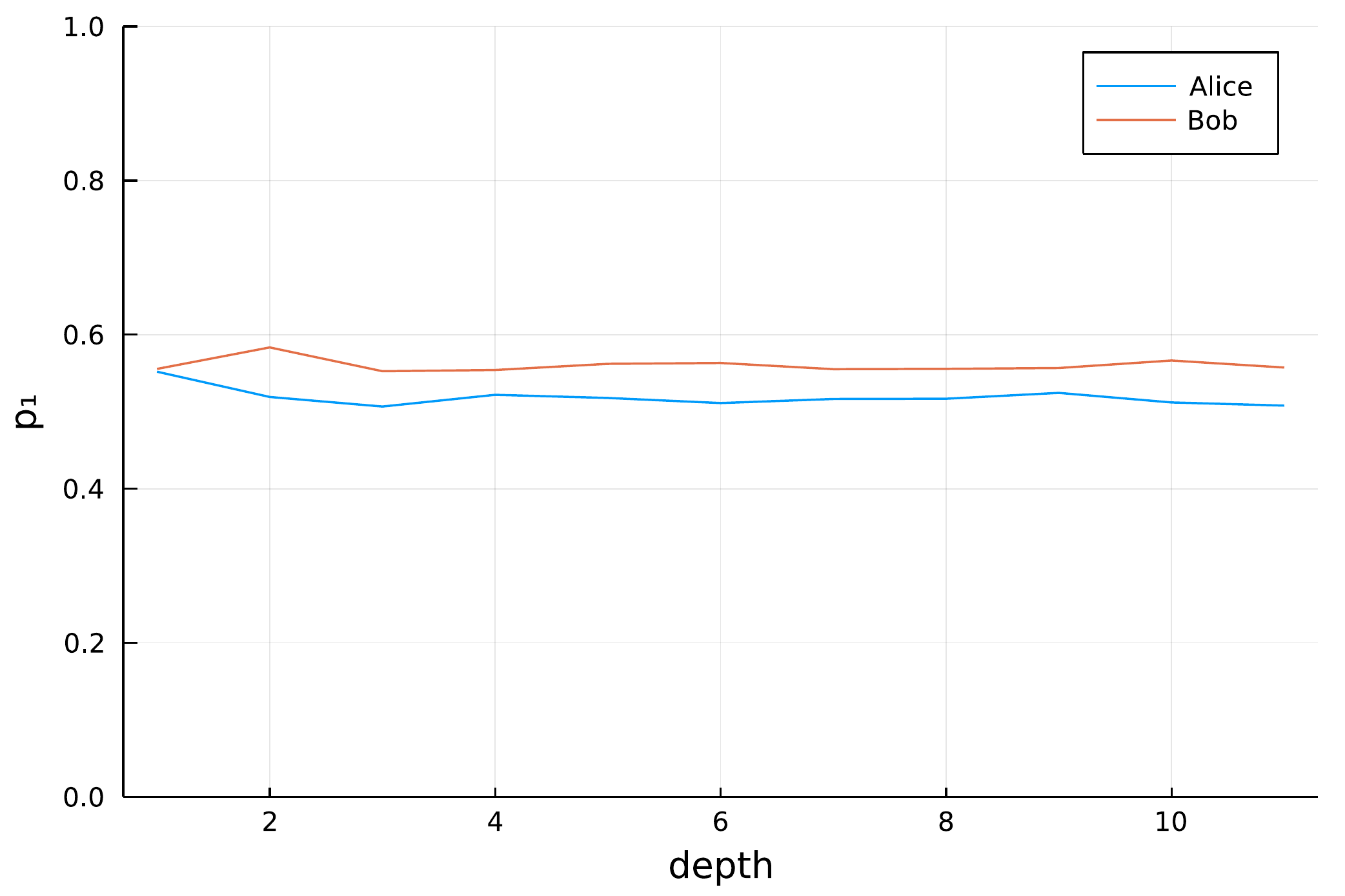}
        \caption{Monte Carlo approximation}
    \end{subfigure}
    \begin{subfigure}{0.495\linewidth}
        \includegraphics[width=\linewidth]{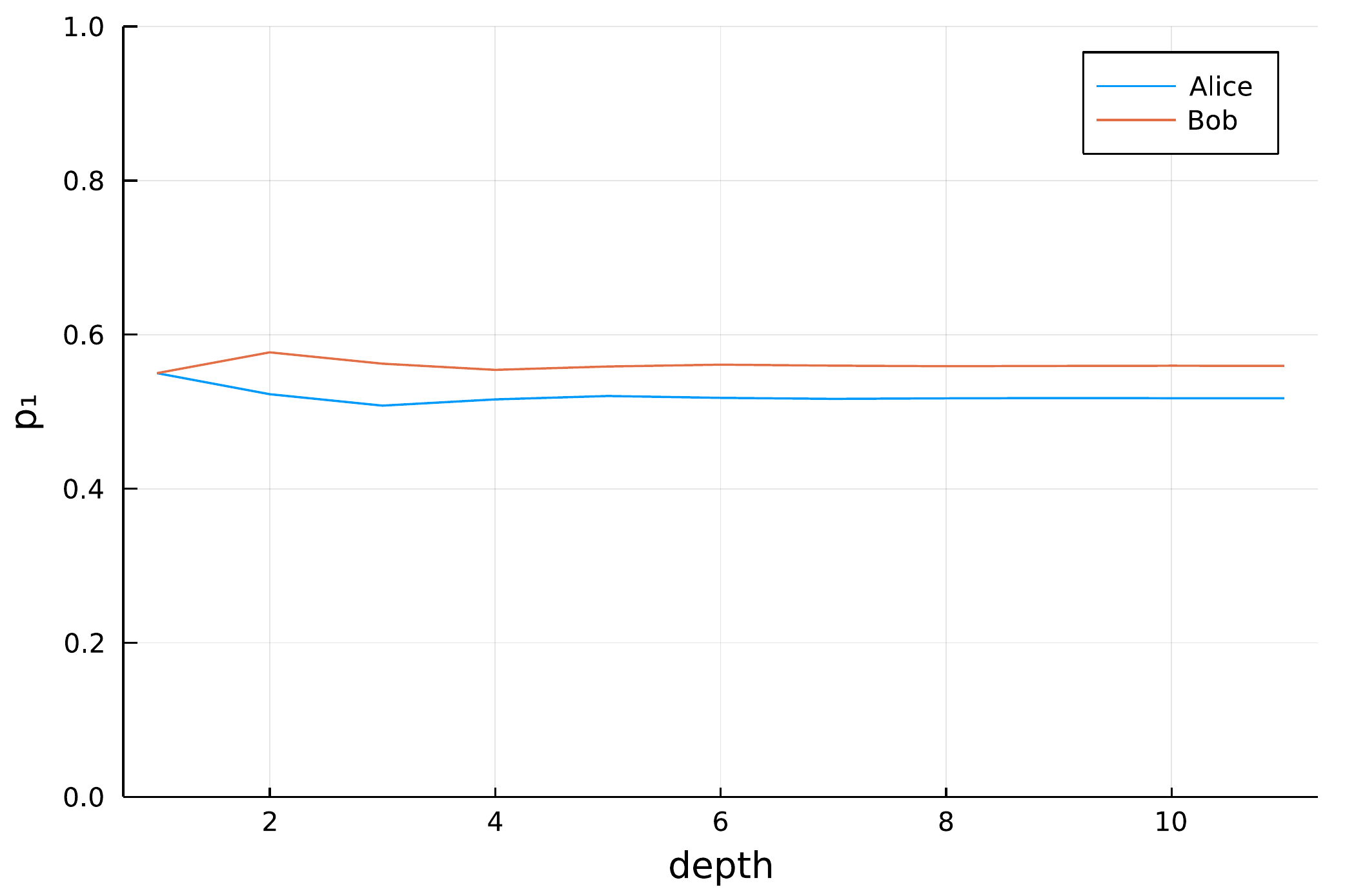}
        \caption{Analytical solution}
    \end{subfigure}
    \caption{Bob chases Alice, mild feelings: $p_{Alice}^1=p_{Bob}^1=0.55$, $p_{Alice}^m=0.25$, $p_{Bob}^m=0.75$}
    \label{fig:ba-0.55-0.25-0.55-0.75}
\end{figure}

Another setting is an instance of pursuit-evasion game: Bob
chases Alice, but Alice avoids Bob. The choice distributions of
both agents depend on how strong their preferences to meet, or
to avoid the meeting, are.
Figure~\ref{fig:ba-0.55-0.25-0.55-0.75}  shows inference results
for the case where Bob's and Alice's feelings are opposite but
mild. As Bob and Alice deliberate deeper and deeper, their
choice distributions eventually converge to stable limits. 

\begin{figure}[h]
    \begin{subfigure}{0.495\linewidth}
        \includegraphics[width=\linewidth]{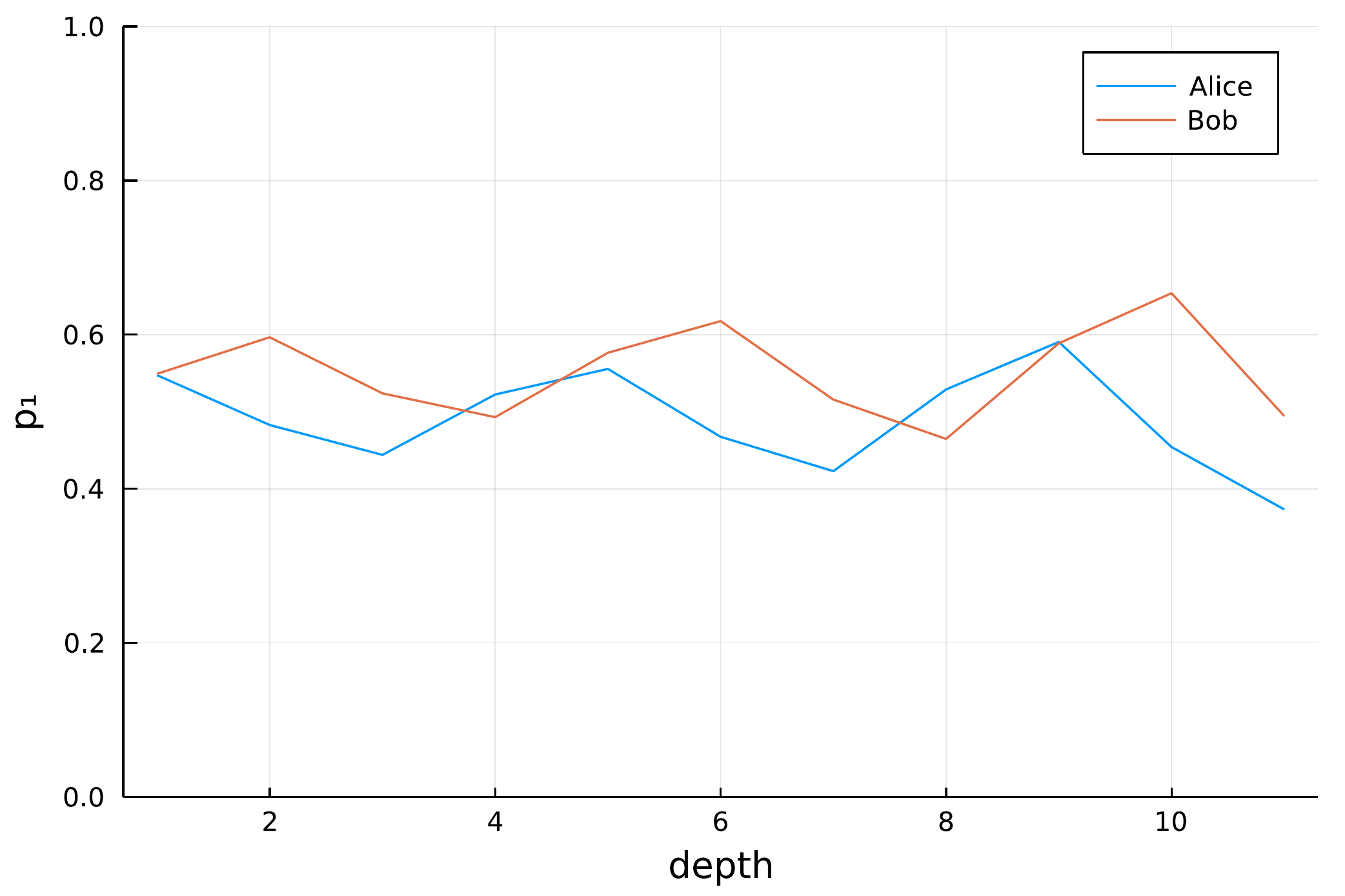}
        \caption{Monte Carlo approximation}
    \end{subfigure}
    \begin{subfigure}{0.495\linewidth}
        \includegraphics[width=\linewidth]{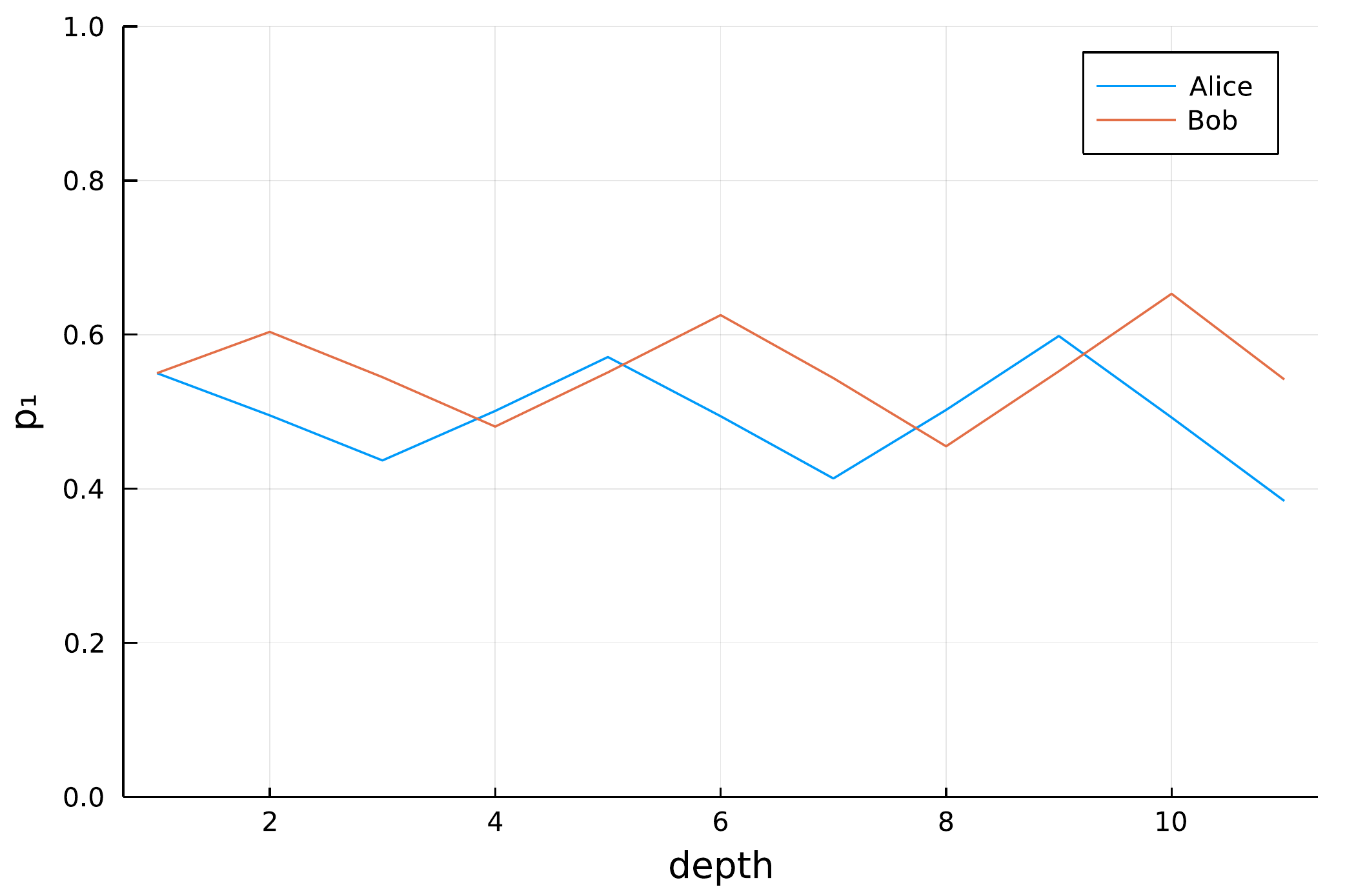}
        \caption{Analytical solution}
    \end{subfigure}
    \caption{Bob chases Alice, strong feelings: $p_{Alice}^1=p_{Bob}^1=0.55$, $p_{Alice}^m=0.1$, $p_{Bob}^m=0.9$}
    \label{fig:ba-0.55-0.1-0.55-0.9}
\end{figure}

However, if Bob's and Alice's feelings are opposite and strong
(Figure~\ref{fig:ba-0.55-0.1-0.55-0.9}),
the agents overthink and act impulsively --- as the deliberation
depth increases, the choice distributions of both agents
alternate between going to the first or to the second bar with
increasing probability.

\subsubsection{Bob and Alice Avoid Each Other}

\begin{figure}[h]
    \begin{subfigure}{0.495\linewidth}
        \includegraphics[width=\linewidth]{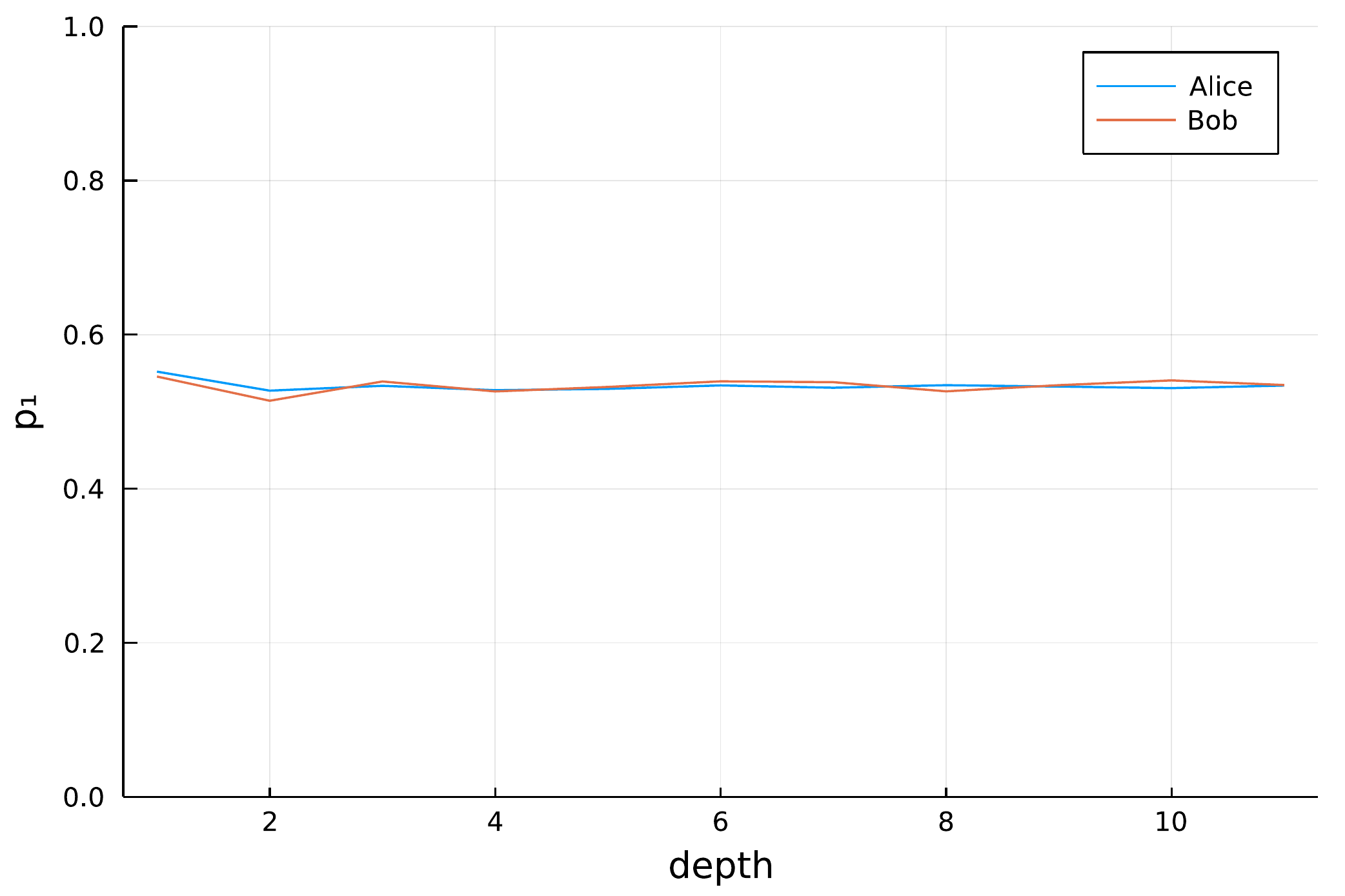}
        \caption{Monte Carlo approximation}
    \end{subfigure}
    \begin{subfigure}{0.495\linewidth}
        \includegraphics[width=\linewidth]{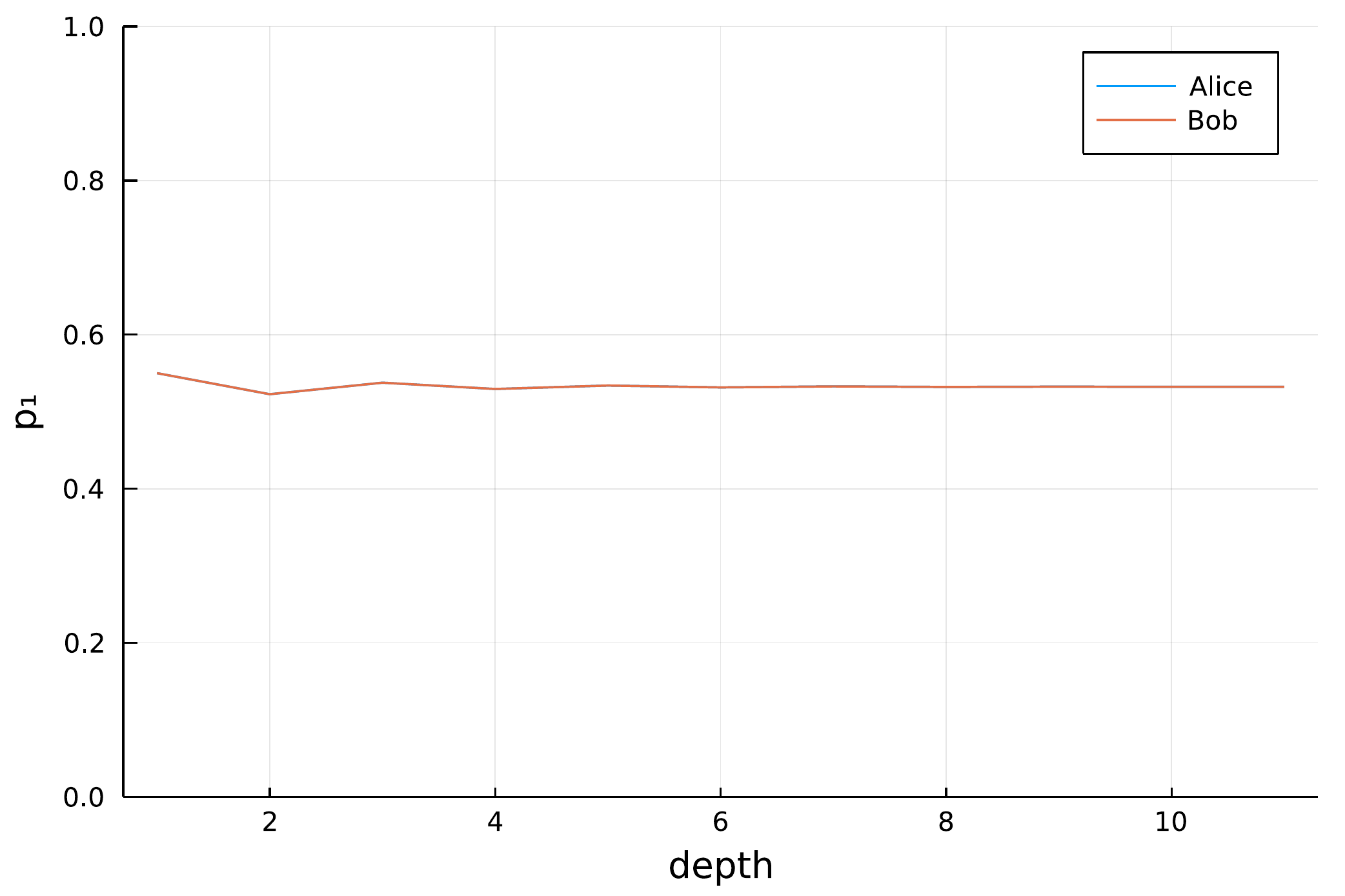}
        \caption{Analytical solution}
    \end{subfigure}
    \caption{Bob and Alice avoid each other, mild feelings: $p_{Alice}^1=p_{Bob}^1=0.55$, $p_{Alice}^m=p_{Bob}^m=0.25$}
    \label{fig:ba-0.55-0.25}
\end{figure}

Further on, we explore the case when Bob and Alice avoid each
other, that is, both of them would rather go to the pub where
they are unlikely to meet the other person. When the preference
towards avoiding is mild (Figure~\ref{fig:ba-0.55-0.25}), Bob
and Alice, with sufficient deliberation depth, approach their
prior preferences, that is they both go to the first bar with
the probability approaching $p_{Alice}^1 = p_{Bob}^1$. Indeed,
rationally this the best they can do to minimize their chances
to meet each other without coordination, while still respecting
their preference of the first bar.

\begin{figure}[h]
    \begin{subfigure}{0.495\linewidth}
        \includegraphics[width=\linewidth]{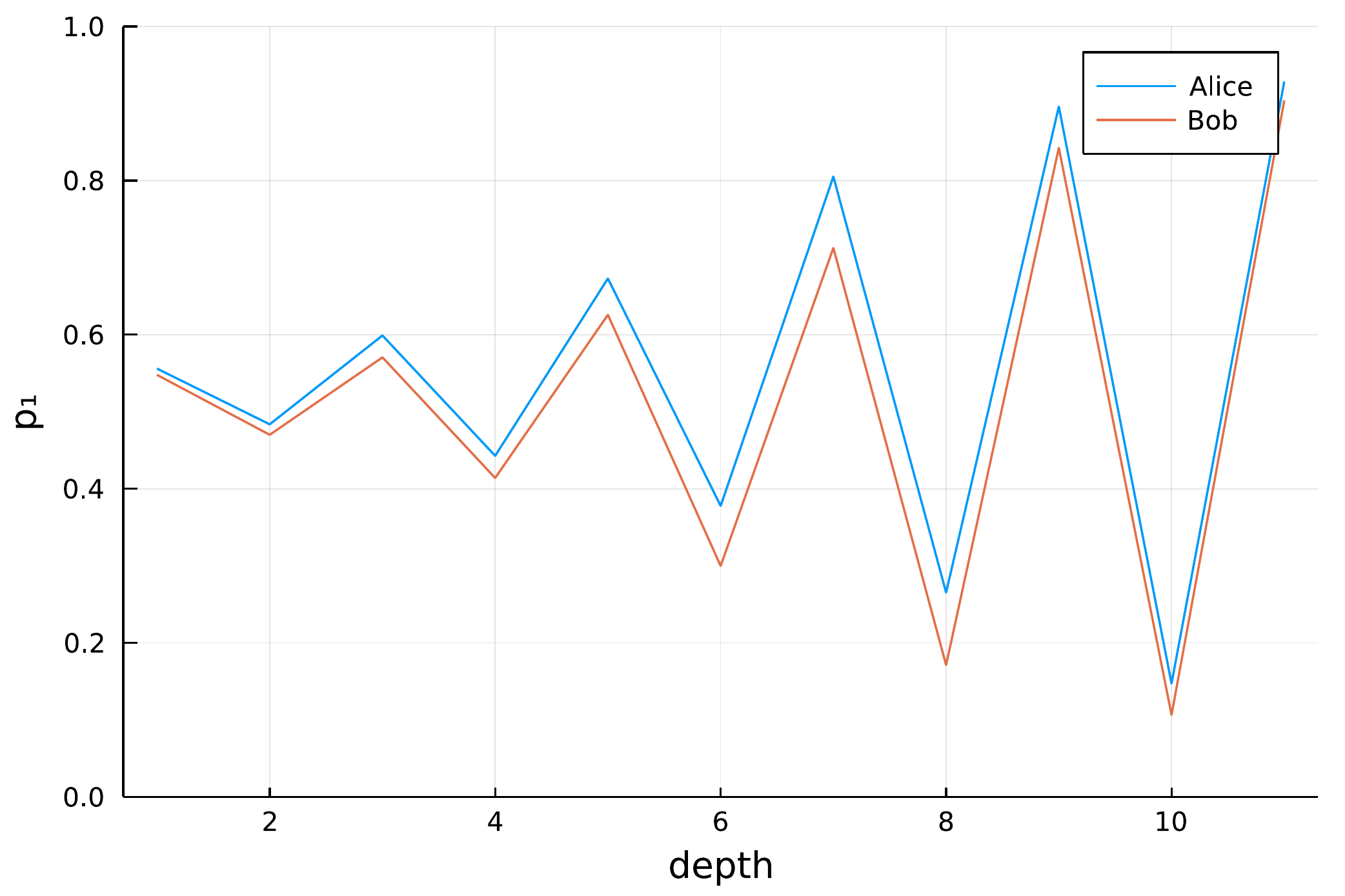}
        \caption{Monte Carlo approximation}
    \end{subfigure}
    \begin{subfigure}{0.495\linewidth}
        \includegraphics[width=\linewidth]{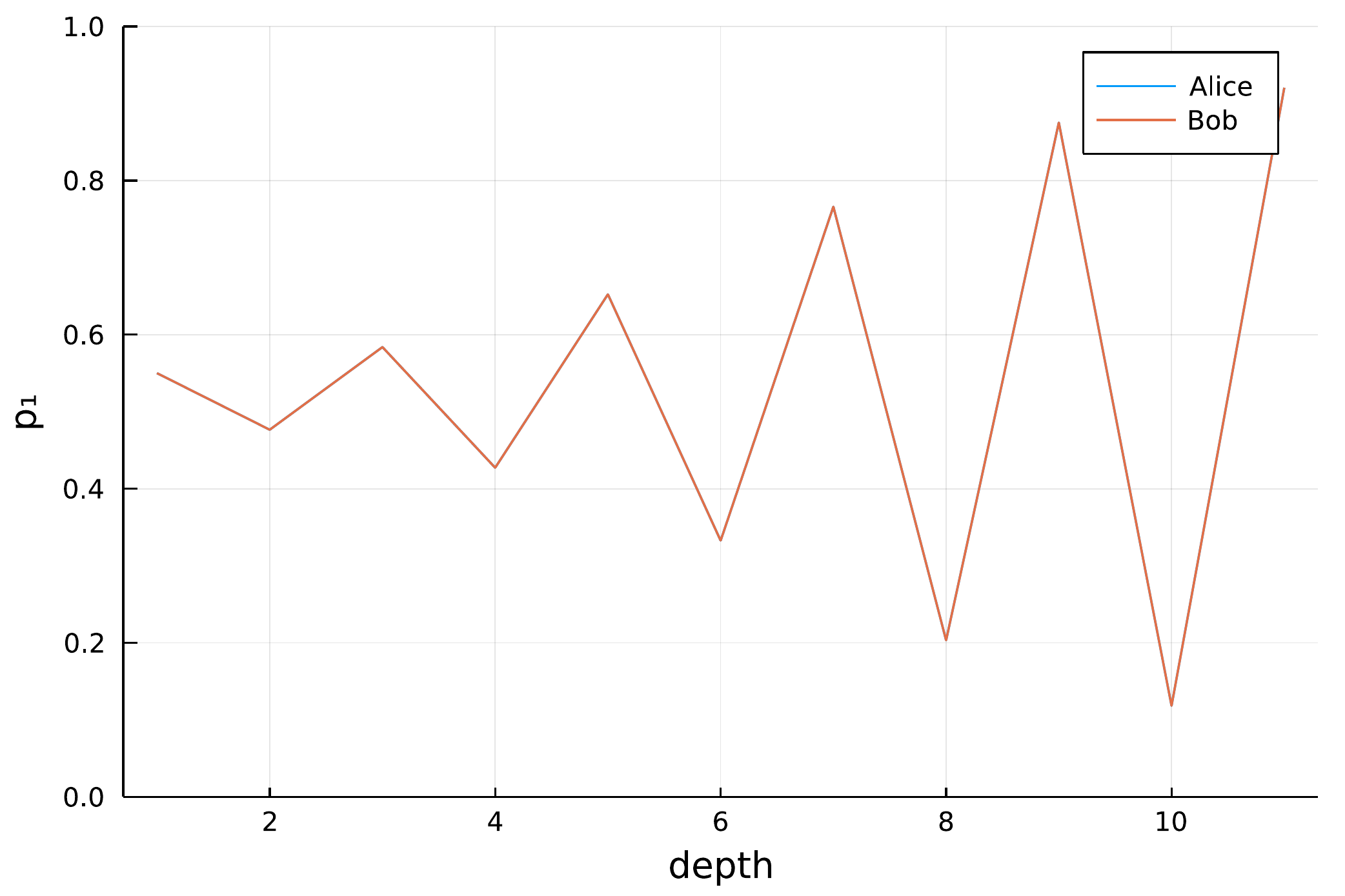}
        \caption{Analytical solution}
    \end{subfigure}
    \caption{Bob and Alice avoid each other, strong feelings: $p_{Alice}^1=p_{Bob}^1=0.55$, $p_{Alice}^m=p_{Bob}^m=0.05$}
    \label{fig:ba-0.55-0.05}
\end{figure}

However, when Bob's and Alice's mutual despisal is too strong
(Figure~\ref{fig:ba-0.55-0.05}), mutual epistemic reasoning does
a poor job --- Bob and Alice switch their increasingly strong
posterior preferences between the first and the second bar with
each level of deliberation depth, similarly to the case of
strong feelings in pursuit-evasion situation. One may argue that 
these outcomes are algorithmically anomalous, or `irrational';
however, they may be also interpreted as a demonstration that
rational behavior is not unconditionally stable, and cannot be
guaranteed for arbitrary combinations of preferences.

\subsubsection{Alice Learns Bob's Preferences}

\begin{figure}[h]
	\includegraphics[width=0.8\linewidth]{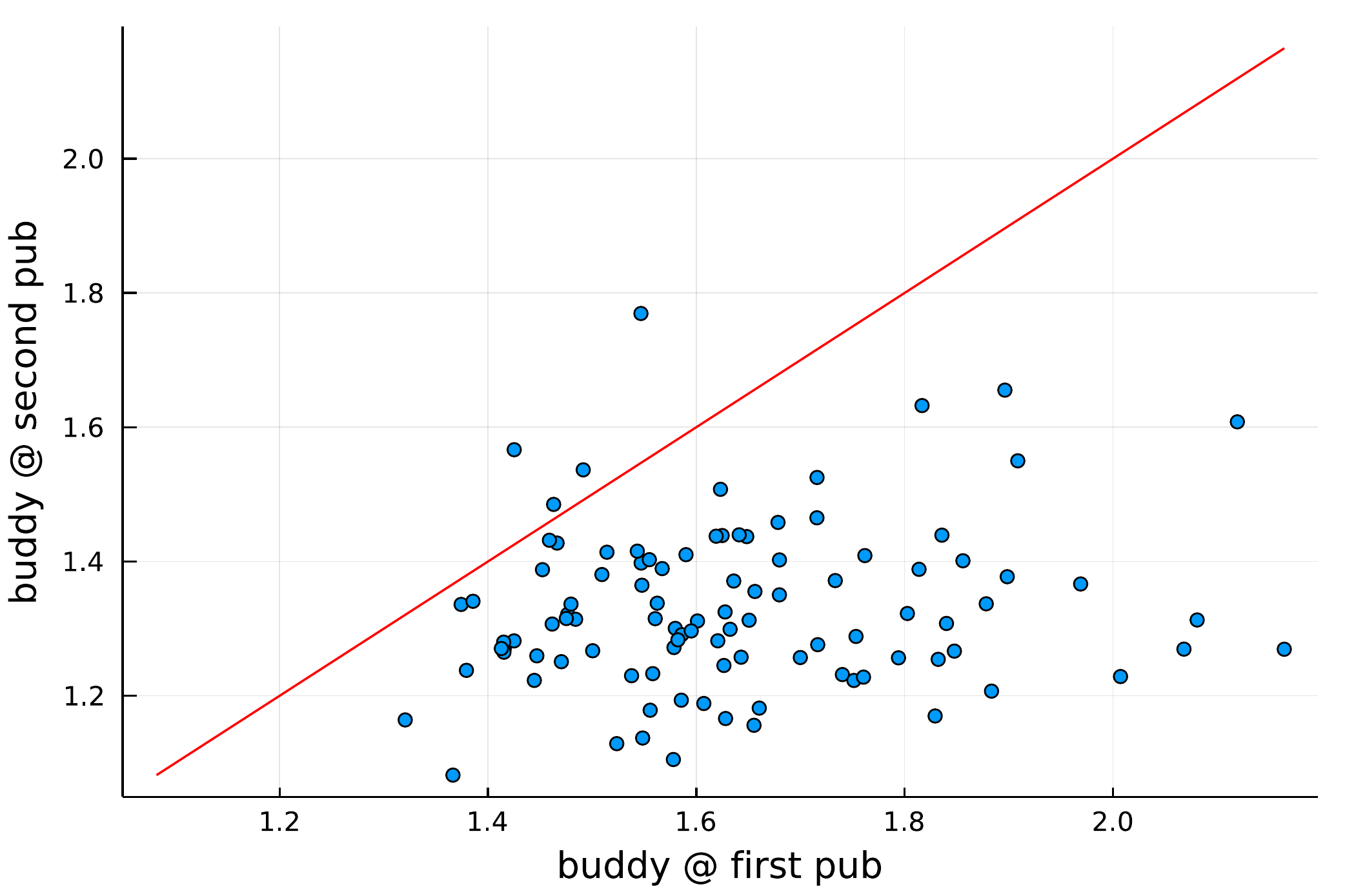}
	\caption{Alice infers Bob preferences: three visits to the
	same bar in a row provide strong evidence about Bob's desire
	to meet Alice.}
	\label{fig:ba-multiround}
\end{figure}

Finally, let us see how Alice can find out whether Bob wants to
meet her by observing Bob's behavior. We use the same model of
Alice for inference, but instead of conditioning the model on
Bob's preferences, we condition the model on the history of
Bob's choices and infer Bob's preference (log-odds) of meeting
Alice. We condition the model on three Bob's choices in a row,
and compare Alice's conclusions for two cases: 
\begin{enumerate}
\item Bob chose the first bar three times in a row; 
\item Bob chose the second bar three times in a row.
\end{enumerate}
To visualize the results, we draw 100 samples from the posterior
distribution of the probability with that Bob would choose to
meet Alice everything else being equal.
Figure~\ref{fig:ba-multiround} shows a two-dimensional scatter
plot where each direction is the marginal distribution of
\textit{log-odds} of Bob willing to meet Alice. Indeed, 96 out
of 100 points are below the diagonal, meaning that Alice can be
very confident that Bob is willing to meet her if she observes
Bob in the first bar 3 times in a row. However, if Bob shows up
three times in a row in the second bar, Alice should conclude
that Bob is avoiding her.

\subsection{The Sailing Problem}

\begin{figure}
\begin{subfigure}{0.495\linewidth}
	\centering
	\includegraphics[width=0.85\linewidth]{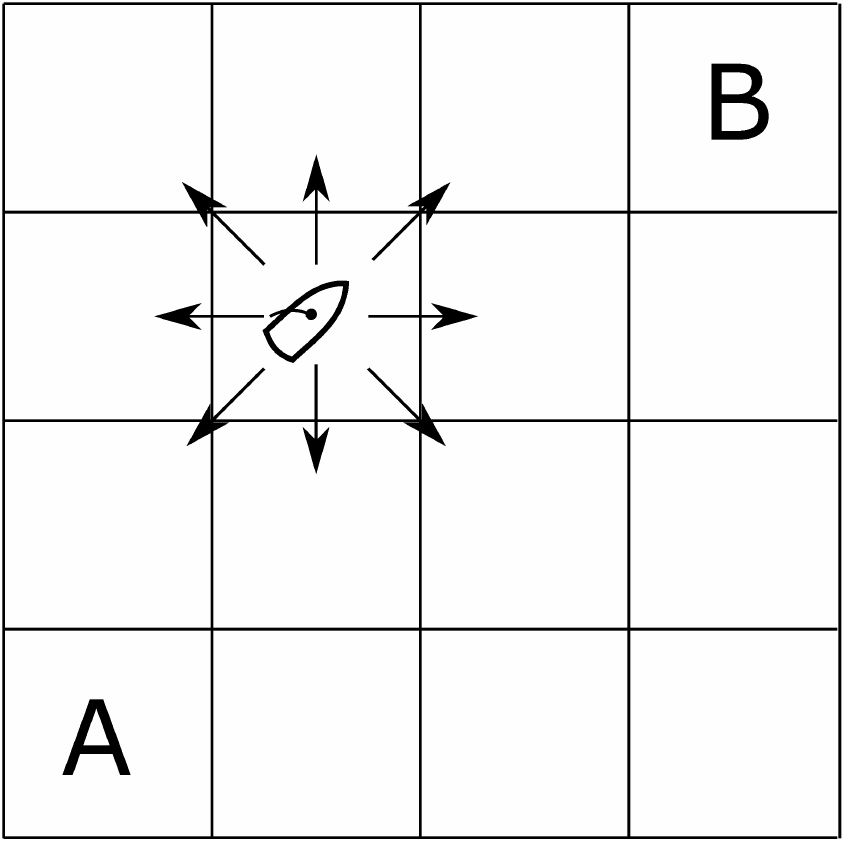}
	\caption{lake}
	\label{fig:sailing-lake}
\end{subfigure}
\begin{subfigure}{0.495\linewidth}
	\centering
	\includegraphics[width=0.85\linewidth]{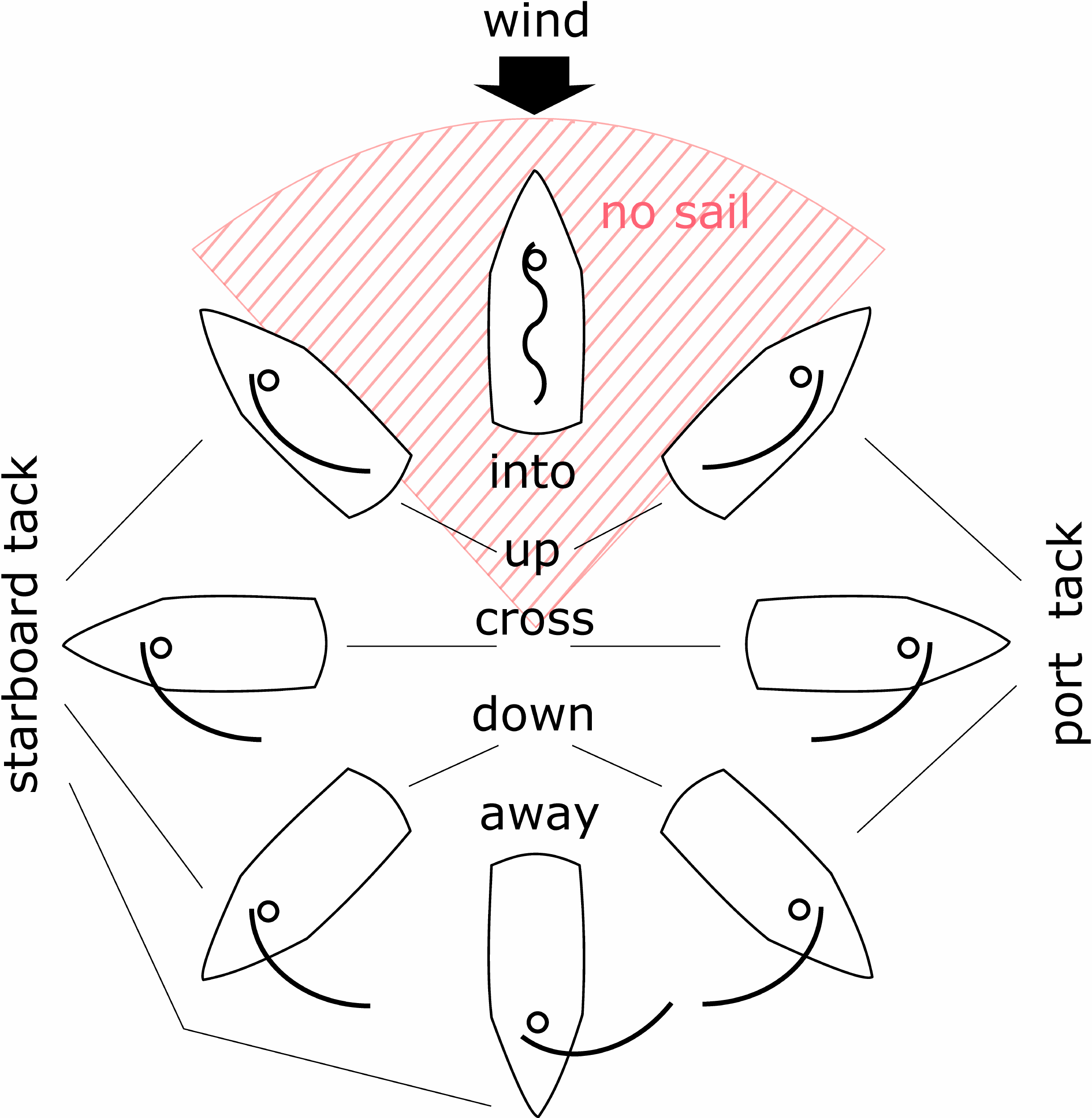}
	\caption{points of sail}
	\label{fig:sailing-points}
	\end{subfigure}
	\caption{The sailing problem}
	\label{fig:sailing}
\end{figure}

The sailing problem (Figure~\ref{fig:sailing}) is a popular
benchmark problem for search and planning. A sailing boat must travel
between the opposite corners A and B of a square lake of a
given size. At each step, the boat can \textit{head} in 8
directions (\textit{legs}) to adjacent squares
(Figure~\ref{fig:sailing-lake}). The
unit distance cost of movement depends on the wind
(Figure~\ref{fig:sailing-points}), which can
also blow in 8 directions. There are five relative
boat and wind directions and associated costs: \textit{into},
\textit{up}, \textit{cross}, \textit{down}, and \textit{away}. 
The cost of sailing into the wind is prohibitively high, upwind
is the highest feasible, and away from the wind is the lowest.
The side of the boat off which the sail is hanging is called the
\textit{tack}, either \textit{port} or \textit{starboard}.
When the angle between the boat and the wind changes sign, the
sail must be \textit{tacked} to the opposite tack,
which incurs an additional \textit{tacking delay} cost.  The
objective is to find a policy that minimizes the expected travel
cost. The wind is assumed to follow a random walk, either
staying the same or switching to an adjacent
direction, with a known probability.

\begin{table}
	\caption{Sailing problem parameters}
	\label{tab:sailing-parameters}
	\setlength\tabcolsep{3pt}
	\centering
	\begin{tabular}{c c c c c c | c c c}
		\multicolumn{6}{c|}{cost} & \multicolumn{3}{c}{wind probability} \\
		into & up & cross & down & away & delay & same & left & right \\ \hline
		$\infty$ & 4  & 3 & 2 & 1 & 4 & 0.4 & 0.3 & 0.3
	\end{tabular}
\end{table}

For any given lake size, there is a non-parameteric stochastic policy
that tabulates the distribution of legs for each
combination of location, tack, and wind. However, such policy does not
generalize well --- if the lake area increases, due to a particularly
rainy year for example, the policy is not applicable to the new
parts of the lake. In this case study, we
infer instead a generalizable parametric policy balancing
between hesitation in anticipation for a better wind and rushing
to the goal at any cost.  The policy chooses a leg with the
log-probability equal to the euclidian distance between
the position after the log and the goal, multiplied by the
policy parameter $\theta$ (the leg directed into the wind is
excluded from choices). The greater the $\theta$, the higher
is the probability that a leg bringing the boat closer to the
goal will be chosen:
\begin{equation}
		\log \Pr(\var{leg}) = \theta \cdot\var{distance}(\var{next-location}, \var{goal}) - \log Z
	\label{eqn:sailing-policy}
\end{equation}
here, $Z$ is the normalization constant ensuring that the
probabilities of all legs sum up to 1. It can be readily
computed but is not required for inference.  We implemented the
model and inference in Infergo~\cite{T19}.

\begin{figure}
	\centering
	\includegraphics[width=0.95\linewidth]{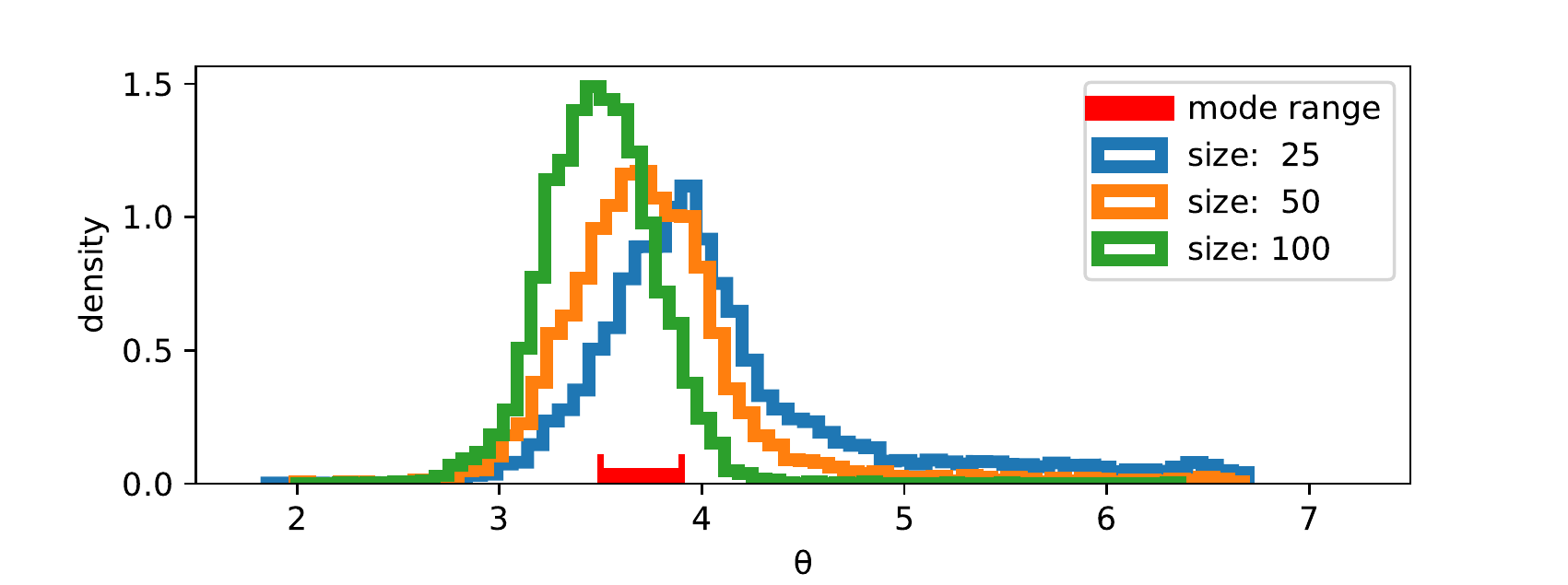}
	\caption{The sailing problem: $\theta$}
	\label{fig:sailing-unit-cost}
\end{figure}

The model turns out to be similar in structure to that of Bob and Alice.
\begin{itemize}
	\item The first agent, Alice, is the boat. The agent chooses
		a path across the lake. 
	\item The second agent, Bob, is the wind. The agent chooses
		a random walk of wind direction along the boat's path.
		The wind is a neutral agent --- it does not try to
		either help or tamper with the boat.
	\item The boat's path, stochastically conditioned on the
		wind, has the probability proportional to the product
		of exponentiated negated leg costs (which are
		interpreted as negated log probabilities of the boat to
		choose each leg regardless of the goal location).
\end{itemize}
Model~\eqref{eqn:sailing} formalizes our setting:
\begin{equation}
	\begin{aligned}
		& \var{wind-history} \sim \var{RandomWalk} \\ \midrule
		& \log \theta \sim \mathrm{Uniform(0, \infty)} \\
		& \var{boat-trajectory} \sim D(\var{wind-history}, \theta) \\
		& \Pr(\var{boat-trajectory}, \theta) \propto \exp(-\var{travel-cost}(\var{boat-trajectory}, \var{wind-history}))
	\end{aligned}
	\label{eqn:sailing}
\end{equation}
The model parameters (cost and wind change
probabilities), same as  in~\cite{KS06,TS12}, are shown in
Table~\ref{tab:sailing-parameters}. A non-informative improper
prior is imposed on $\theta$.  We fit the model using pseudo-marginal
Metropolis-Hastings~\cite{AR09} and used $10\,000$ samples to
approximate the posterior.
Figure~\ref{fig:sailing-unit-cost} shows the posterior distribution
of the unit cost.
Table~\ref{tab:sailing-travel-cost} shows the
expected travel costs, with the expectations estimated both over
the unit cost and the wind. The inferred travel costs are
compared to the travel costs of the `optimal' policy
and of the greedy policy, according to which the boat always
heads in the direction of the steepest decrease of the distance
to the goal. One can see that the inferred policy attains an
expected travel cost lying between the greedy policy and the
`optimal' policy, as one would anticipate.

\begin{table}
	\centering{}
	\caption{The sailing problem: travel cost}
	\label{tab:sailing-travel-cost}
	\begin{tabular}{r|c|c|c}
		 & 25 & 50 & 100 \\ \hline
		Inferred & 105 &  209 & 413 \\
		Greedy & 107 & 215 & 430 \\
		Optimal & 103 & 206 & 406 
	\end{tabular}
\end{table}

\section{Related Work}

Possibility and importance of casting reinforcement learning as
probabilistic inference are well understood in AI
research~\cite{TS06,BA09}.  \cite{KZT11} explore planning as
inference in the case of multiple agents, exploring special
cases in which efficient scalable inference is possible.

Although not strictly in the field of reinforcement learning,
\cite{K07} lays out stochastic control theory and shows the
connection between Boltzmann distribution and expected utility
maximization.  \cite{WGR+11} demonstrates usefulness and
applicability of probabilistic programming for planning as
inference on deterministic planning domains. \cite{MPT+16} apply
probabilistic programming to stochastic domains, using a custom
policy search algorithm.

\cite{SG14} explore multiagent settings using probabilistic
programming, implementing theory of mind through mutually
recursive nested condition of models. \cite{SMW18} apply theory
of mind using probabilistic programming to an elaborated
pursuit-evasion problem.

\cite{TZR+21} introduces stochastic conditioning, a Bayesian
inference formalism necessary to cast reinforcement learning as
a purely probabilistic Bayesian inference problem. In one of the
case studies, \cite{TZR+21} provide an early indication that
planning as inference can be implemented through stochastic
conditioning.

\section{Discussion}

We demonstrated that reinforcement learning, that is, agent's
reasoning about preferred future behavior, can be formulated as
Bayesian inference. Probability distributions can be used to
specify all aspects of uncertainty or stochasticity, including
agent preferences. Rewards can be interpreted as log-odds of
stochastic preferences and do not need to be explicitly
introduced.  Planning algorithms maximizing the expected utility
find \textit{maximum a posteriori} characterization of the
rational policy distribution.

\bibliography{refs}
\bibliographystyle{alpha}

\end{document}